\documentclass{article}

\usepackage[preprint]{neurips_2020}
\usepackage{float}
\usepackage{graphicx,wrapfig}
\usepackage[utf8]{inputenc}
\usepackage[T1]{fontenc}
\usepackage[english]{babel}
\usepackage{amsthm}
\usepackage{hyperref}      
\usepackage{url}          
\usepackage{booktabs}     
\usepackage{amsfonts}       
\usepackage{nicefrac}       
\usepackage{microtype}      
\usepackage{mathtools}
\usepackage{tikz}
\usepackage{graphicx}
\usepackage{pgfplots}
\usepackage{multirow}
\usepackage{subcaption}
\usepgfplotslibrary{groupplots}
\usepackage{natbib}
\usepackage{bm,bbm}
\usepackage[ruled,vlined]{algorithm2e}
\newcommand\R{\mathbb{R}}

\newtheorem{remark}{Remark}[section]
\usepackage{tikz}
\usetikzlibrary{positioning}
\usetikzlibrary{shapes.geometric, arrows}
\tikzstyle{Relu} = [isosceles triangle, draw=red,rotate=90, minimum size =0.5cm,fill=blue!20]
\tikzstyle{HRelu} = [isosceles triangle, draw=red,dashed,rotate=90, minimum size =0.5cm,fill=blue!20]
\tikzstyle{linear} = [circle, minimum size =0.5cm, draw=red, fill=red!30]
\tikzstyle{linear1} = [circle, minimum size =0.8cm, draw=red, fill=red!30]
\tikzstyle{box} = [rectangle, minimum width =1cm, minimum height =0.5cm, draw=black,dotted, fill=green!30]
\tikzstyle{box_weight} = [rectangle, minimum width =1cm, minimum height =0.5cm, draw=red, fill=red!30]
\tikzstyle{input} = [rectangle, minimum width =1cm, minimum height =0.5cm, draw=white, fill=white]
\tikzstyle{arrow} = [thick,->,>=stealth]

\tikzstyle{bias} = [circle, draw=red, fill=red!30,minimum size = 0.2cm, inner sep=0pt]
\tikzstyle{tool} = [trapezium, trapezium left angle=70, trapezium right angle=110, minimum width=3cm, minimum height=1cm, text centered, text width=2cm, draw=black, fill=blue!30]
\tikzstyle{process} = [rectangle, 
minimum width=3cm, 
minimum height=1cm, 
text centered, 
text width=3cm, 
draw=black, 
fill=orange!30]
\tikzstyle{inputs} = [diamond, 
minimum width=3cm, 
minimum height=1cm, 
text centered, 
draw=black, 
fill=green!30]
\pgfplotsset{compat=1.16}

\title{Non-Parametric Estimation of  Multi-dimensional Marked Hawkes Processes }

\author{
	Sobin Joseph\\
	Department of Management Studies\\
	Indian Institute of Science\\
	Bangalore 560012\\
	\texttt{sobinjoseph@iisc.ac.in}\\
	\And
	Shashi Jain\\
	Department of Management Studies\\
	Indian Institute of Science\\
	Bangalore 560012\\
	\texttt{shashijain@iisc.ac.in} \\
}
\begin{document}
\newcommand{\boldcal}[1]{\boldsymbol{{\mathcal{#1}}}}
	\maketitle
	\begin{abstract}
		
An extension of the Hawkes process, the Marked Hawkes process distinguishes itself by featuring variable jump size across each event, in contrast to the constant jump size observed in a Hawkes process without marks. While extensive literature has been dedicated to the non-parametric estimation of both the linear and non-linear Hawkes process, there remains a significant gap in the literature regarding the marked Hawkes process. In response to this, we propose a methodology for estimating the conditional intensity of the marked Hawkes process. We introduce two distinct models: \textit{Shallow Neural Hawkes with marks}- for Hawkes processes with excitatory kernels and \textit{Neural Network for Non-Linear Hawkes with Marks}- for non-linear Hawkes processes. Both these approaches take the past arrival times and their corresponding marks as the input to obtain the arrival intensity. This approach is entirely non-parametric, preserving the interpretability associated with the marked Hawkes process. To validate the efficacy of our method, we subject the method to synthetic datasets with known ground truth. Additionally, we apply our method to model cryptocurrency order book data, demonstrating its applicability to real-world scenarios.  
	\end{abstract}


	\section{Introduction}
	\label{Sec:intro}

\begin{figure}[ht]
     \centering
     \begin{subfigure}[b]{0.45\textwidth}
        \centering
        \includegraphics[width=\linewidth]{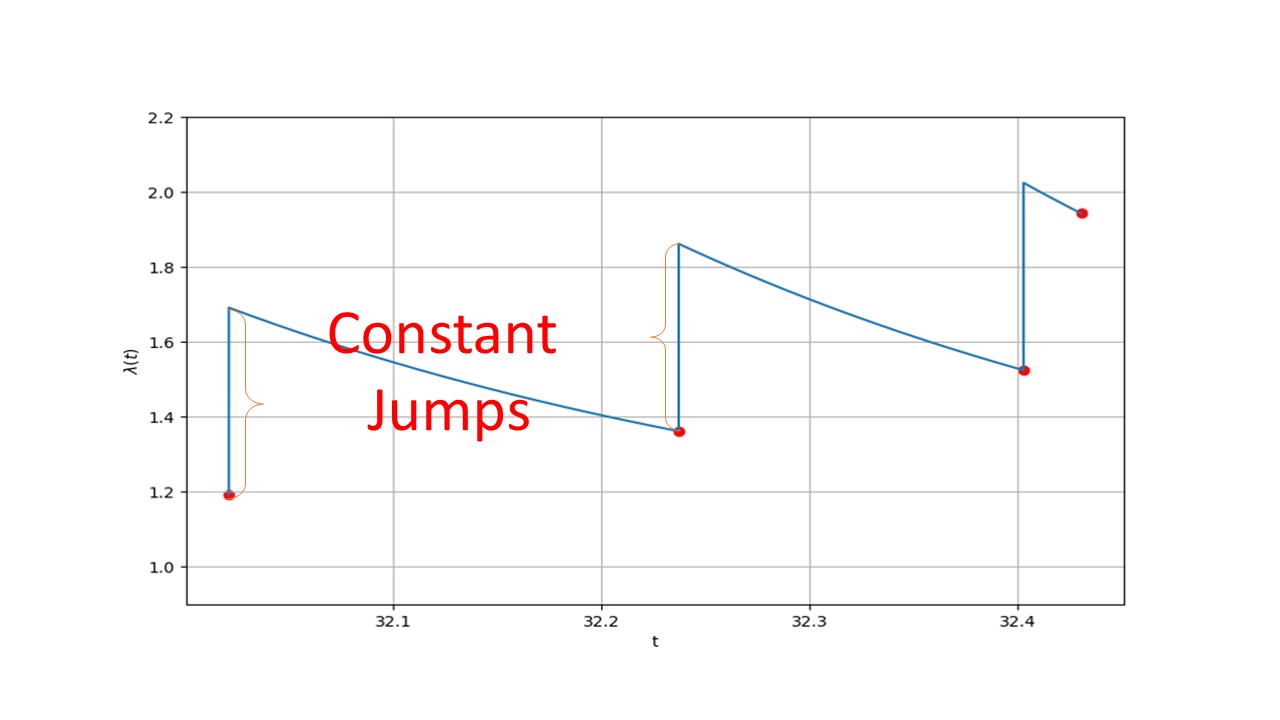}
        \caption{Hawkes Process}\label{Fig:hp}
     \end{subfigure}
     \hfill
     \begin{subfigure}[b]{0.45\textwidth}
         \centering
        \includegraphics[width=\linewidth]{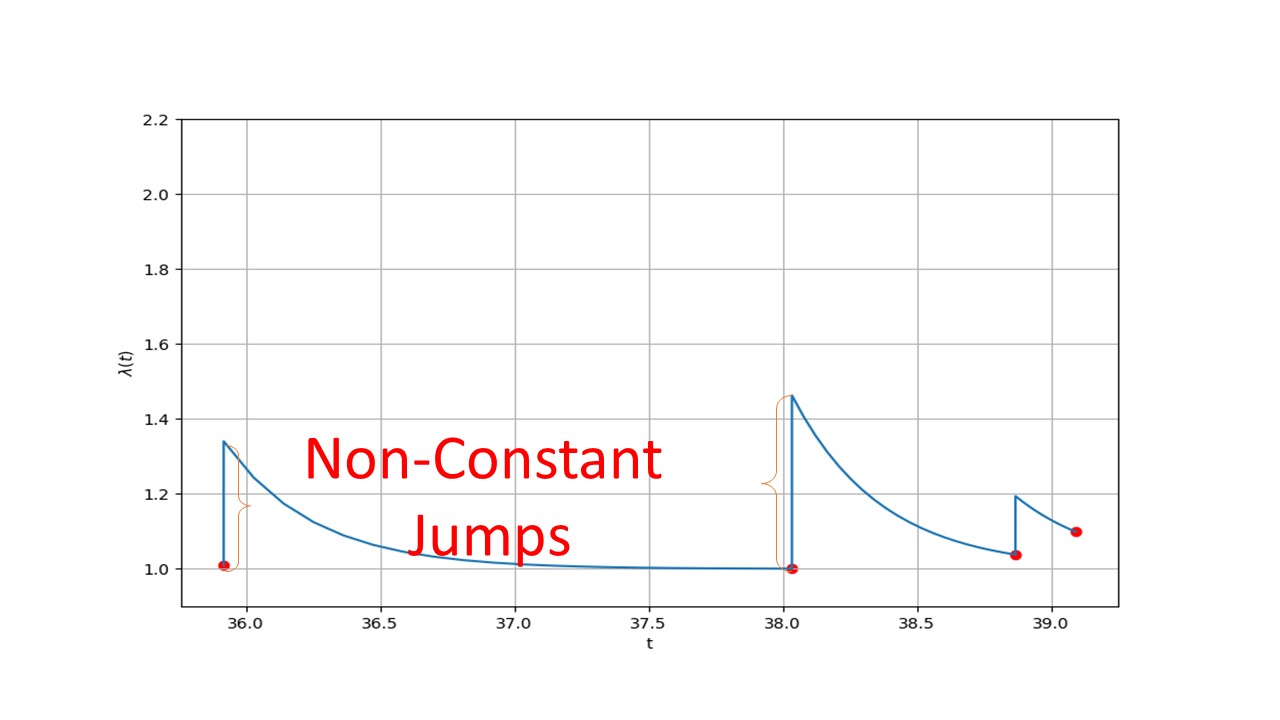}
        \caption{Marked Hawkes Process}\label{Fig:mhp}
    \end{subfigure}
    \caption{Comparison between Hawkes process and Marked Hawkes process}
    \label{Fig:HP&MHP}
    \end{figure}

Hawkes process proposed by A.G. Hawkes is a self or mutually exciting multivariate point process with the intensity function dependent on time steps of past events \cite{Hawkes1971spectra}.  Marked Hawkes process is an extended version of the Hawkes process, where the conditional intensity function considers the past timestamps of events and the associated mark with each timestamp. These marks can represent various factors, aside from event time, that directly affect the intensity of the events. Consider the scenario of market order arrivals in a exchange; the time of the next market order is not only determined by the history of order arrivals but is also influenced by the history of past order volumes. This illustrates a marked Hawkes process where, in an arrival process, the occurrence of the next event depends on both the arrival history and the mark history (in this example, order volume serves as the mark component).  The marked Hawkes process has found applications in diverse fields such as seismology where magnitude and position of the earthquake are the mark variables (\cite{ogata1998space}, \cite{zhuang2004analyzing}, and \cite{fox2016spatially}), in social networks where the number of followers serves as a mark category (\cite{kobayashi2016tideh}, and \cite{chen2018marked}), in finance which has volume (\cite{chavez2012high} and \cite{fauth2012modeling}) and the occurrence of extreme events (\cite{embrechts2011multivariate}, and \cite{stindl2019modeling})  as the mark variable and, in criminology with fraud transactions as marks (\cite{yuan2019multivariate}, and \cite{narayanan2022point}). In a Hawkes process without marks, when an event occurs, it can excite or inhibit more events. The size of each excitation or decay, referred to as the jump size, remains constant across all events. In contrast, in a marked Hawkes process, an event's impact on future events is characterized by varying jump sizes. Figure \ref{Fig:HP&MHP}  illustrates the distinction between a Hawkes process and a marked Hawkes through an intensity-time plot. In this plot, we observe a constant intensity jump for Hawkes and a variable jump for the marked Hawkes process. 

 For a marked Hawkes process, similar to a Hawkes process, the intensity function combines exogenous background rate and endogenous kernels. This combination can take linear or non-linear forms. A Hawkes process, which models the self-excitability of events based on their past occurrences, can be limited in explaining dependencies related to the characteristics of the event, such as volume, price, magnitude, number of followers or any other relevant features.  Consider the earlier example of market order arrivals in an exchange; even though a Hawkes process without marks can explain the dependency of an order arrival with its past arrival, it won't be able to explain the effects of order volume on market order evolution. But considering it as a marked Hawkes process allows the model to capture not only the temporal dependencies between successive orders but also the impact of the volume of past orders on the evolution of the market. This is essential in understanding how the arrival of market orders is influenced by past order volumes.  Take another example of earthquake aftershocks: the magnitude of an earthquake serves as a crucial mark that influences the likelihood of further aftershocks. A marked Hawkes process enables the modelling of the arrival intensity of aftershocks, considering both temporal dependencies between the aftershocks and the influence of their magnitudes.
 
 The kernels in a marked Hawkes process intensity function play a pivotal role in capturing the temporal and mark dependencies. They quantify the influence of past events and their marks on the likelihood of future events. Thus understanding the form of these kernels is crucial for accurate modelling and estimation in the marked Hawkes process. 

Hence here, we propose a novel feed-forward neural network-based approach to model the marked Hawkes process kernel function. Our method utilizes a 2-layered feed-forward neural network to capture the Hawkes kernel using the history of event times and their associated marks.  The network design includes a hidden layer and an output layer for approximating the kernel function that receives both the historical data of past arrivals and their corresponding marks as  the input. This modelling approach extends the neural network-based models, specifically the Shallow Neural Hawkes (SNH) (\cite{joseph2022shallow}) and the Neural Network for Non-Linear Hawkes (NNNH) (\cite{joseph2023neural}), which were initially designed for Hawkes processes without marks. 

\subsection{Literature Review of Marked Hawkes Process}
While there has been considerable research on estimating the Hawkes kernels and base intensity, the literature focused specifically on the marked Hawkes process is relatively scarce. Most of the available studies on estimating the marked Hawkes process primarily revolve around parametric assumptions, with only a few exceptions exploring non-parametric methods.

The early studies of the marked Hawkes process primarily focused on predicting earthquake aftershocks, known as ETAS (Epidemic-Type Aftershock Sequence). The ETAS model was developed in \cite{ogata1988statistical} and \cite{ogata1999seismicity}, considering past earthquake event times as the temporal variable and earthquake magnitude as the mark variable. Subsequently, \cite{zhuang2004analyzing} modified this approach to incorporate the event location as an additional spatial mark factor. It's worth noting that both methods rely on parametric assumptions for the kernel and mark function. \cite{embrechts2011multivariate} leans towards parametric assumptions, applying the marked Hawkes process to model extreme price moves in stock market index data.  Similarly, \cite{chen2018marked} adopts a parametric marked Hawkes process to model retweet cascading on  Twitter data. 

Transitioning to non-parametric approaches for the marked Hawkes process, \cite{bacry2016first} uses the Wiener-Hopf integral, while \cite{fox2016spatially} employs the EM algorithm for estimating the marked Hawkes kernel and base intensity. Although both models are non-parametric, they model the temporal and mark part of the kernel as distinct functions with no interdependence. Furthermore, several non-parametric techniques based on neural networks have been proposed. For instance, \cite{mei2017neural}, \cite{du2016recurrent} and \cite{shchur2019intensity} utilize recurrent neural network (RNN)-based models with maximum likelihood estimation (MLE) as the loss function to estimate the marked Hawkes process. Another approach, as proposed by \cite{xiao2017wasserstein}, adopts the RNN-based Wasserstein generative adversarial network (WGAN) to learn the marked Hawkes process. However, it's worth noting that RNNs may face challenges in capturing extended dependencies and encounter issues like gradient vanishing and gradient explosion, as highlighted by \cite{pascanu2013difficulty}. Introducing a different perspective, \cite{zuo2020transformer} presents the transformer Hawkes, which utilizes self-attention modules combined with feed-forward networks to learn the marked Hawkes process. \cite{fabre2024neural} utilize the physics-informed neural network (PINN) to estimate the kernels of a marked Hawkes process. This model incorporates the first and second-order characteristics equation of \cite{bacry2016first} to facilitate the training of the neural network. 

\subsection{Contribution and Organization of the paper}

The aforementioned methods employ neural networks to directly model the conditional intensity function, making it challenging to deduce the causal relationships between each event and its historical context. Even though neural network-based methods have conducted joint analyses of marks and time, those approaches typically handle discrete marks (marks in $\mathcal{Z}^+$).   In the approaches that consider marks as continuous, such as \cite{fabre2024neural}, the marked Hawkes kernel function is taken as a decoupled function of time and mark.

Thus, as our first contribution,  we propose a novel approach based on neural networks, which can approximate the marked Hawkes process kernel instead of the intensity function. This allows for inferring causal relationships between events across each dimension using the estimated kernels. The two-layered feed-forward network accommodates continuous marks, enabling a more comprehensive modelling of the marked Hawkes process. The parameters of the model are estimated by maximizing the log-likelihood function. Since the log-likelihood function is non-convex for a Hawkes process, even for a parametric setting (\cite{joseph2022shallow}), we use the stochastic gradient descent (SGD) method to derive the network parameters. The model represents a novel approach capable of jointly modelling a marked Hawkes kernel's temporal and mark dependency.

The application of the proposed method to the high-frequency cryptocurrency trading data is our next contribution. The dataset comprises the arrival times and their associated volumes for market orders in Bitcoin-US Dollar and Ethereum-US Dollar pairs. From the proposed method, we obtain the underlying kernels, providing insights into the relationship of an event concerning both the history of time and the mark. This improves predictions of future arrivals and uncovers connections between arrival patterns, historical data, and associated order volumes. This insight can provide a clearer understanding of the underlying dynamics of the market's microstructure.

The remainder of the paper is structured as follows. Section \ref{Sec:prelim} provides the definition of the marked Hawkes process and its associated log-likelihood function. In Section \ref{Sec:model}, we establish the neural network models of the SNH with marks and the NNNH with marks. Section \ref{Sec:results} is dedicated in validating the performance of our proposed model, with evaluations conducted on simulated datasets where the ground truth is known. Furthermore, we apply the estimation method to a real-life cryptocurrency market order arrivals dataset, considering order volume as the mark component. Finally, Section \ref{Sec:conclusion} presents a concise conclusion of the paper.
	
\section{Preliminary Definitions}\label{Sec:prelim}

\paragraph{Definition of the Marked Hawkes Process:}

As per the definition provided by \cite{daley2007introduction},  a marked point process $N(\mathcal{X} \times \mathcal{M})$, with event time in the space $\mathcal{X}$ and marks in the space $\mathcal{M}$, is a point process $\{(t_n,m_n)\}_{n \ge 1}$ on $\mathcal{X} \times \mathcal{M}$ with the additional property that ground process $N_g(.)$ ($N_g$ constitutes the collection of the event times $\{t_n\}$) is itself a point process; i.e. for a bounded set $A \in \mathbb{B}_{\mathcal{X}}$ (where $\mathbb{B}_{\mathcal{X}}$ is the Borel set of the space $\mathcal{X}$), $N_g(A)= N(A \times M) < \infty$.

A D-dimensional marked point process defined as $\left\{N_1(\mathcal{X}\times \mathcal{M}), \ldots, N_D(\mathcal{X}\times \mathcal{M})\right\}$, where each $N_d(\mathcal{X} \times \mathcal{M}) = N^g_d(t)$ for $d = 1,\ldots, D$ represents a marked point process. We introduce $\boldsymbol{\mathcal{H}} \equiv {\mathcal{H}_t: t \ge 0 }$ as the internal history, where $\mathcal{H}_t$ is the history of the process up to and including time $t$, and $\mathcal{H}_{t-}$ is the history of the process up to but not including time $t.$ For the given marked point process $N:= \left\{N_d(\mathcal{X} \times \mathcal{M})\right\}_{d=1}^D$, the mark space $\mathcal{M}$ can take various forms: it can be discrete, as in the case where marks represent the number of fraudulent transactions (\cite{narayanan2022point} or the number of Twitter followers (\cite{kobayashi2016tideh}), positive continuous when marks represent energy (\cite{ogata1998space}), volume (\cite{chavez2012high}) or price (\cite{lee2017marked}), or even multidimensional Euclidean space for space-time processes. In this paper, without loss of generality, we explore examples where $\mathcal{M}$ is positively continuous, denoted as $\mathcal{M} \in \mathbb{R}^{+}$, and $\mathcal{X} \in [0, T).$

 The $d$th dimensional conditional intensity of the marked point process $N$ defined on $[0,T) \times \mathcal{M}$, with respect to its internal history $\boldsymbol{\mathcal{H}}$ represented by  $\lambda_d(t,m)$ is given by,

	\begin{equation}
		\label{Eq:PPIntensity}
		\lambda_d(t,m) dt\,dm \approx \mathbb{E}\left[N_d(dt \times dm)|\mathcal{H}_{t-}\right].
	\end{equation}
 The given intensity $\lambda_d(t,m)$ can also be represented as,
	\begin{equation}
		\label{Eq:PPIntensityDen}
		\lambda_d(t,m) =\lambda^g_d(t|\mathcal{H}_{t-}) f_d(m|t,\mathcal{H}_{t-}),
	\end{equation}
where $\lambda^g_d(t|\mathcal{H}_{t-}),$ is the conditional intensity of the ground process (from this point onwards, for ease of notation we will denote $\lambda^g_d(t|\mathcal{H}_{t-})$ as $\lambda^g_d(t)$), and $f_d(m|t,\mathcal{H}_{t-})$ is the conditional density of a mark at $t$ given $\mathcal{H}_{t-}.$ This emphasizes the significance of the distribution of marks in defining the Hawkes intensity function. Consider the following definitions from  \cite{daley2007introduction}:
	
\begin{itemize}
\item \emph{Independent marks:}
$N_d := N_d([0,T)\times \mathcal{M})$ has independent marks if, for a given set of event times $\{t^d_n\}_{n \ge 1},$ the marks $\{m^d_n\}_{n \ge 1}$ are mutually independent random variables such that the distribution of $m^d_n$ depends only on the corresponding event time $t^d_n$.
\item \emph{Unpredictable marks:} 
$N_d$ has unpredictable marks if the distribution of mark at $t$ is independent of its history $\mathcal{H}_{t-}$.
\end{itemize}

In the context of a marked counting process, the independence of marks implies that both marks and the event times are independent of each other, i.e. the marks don't influence event times and vice-versa. Additionally, unpredictable marks occur when the distribution of marks can influence the subsequent evolution of the event times, but event times don't influence the distribution of marks. This work's proposed models are suitable for independent and unpredictable marks. For both the independent and unpredictable marks case, the conditional mark density is i.i.d. with $f_d(m|t,\mathcal{H}_{t-})= f_d(m).$
	
 The conditional intensity, $\lambda_d(t,m),$ for the $d$-th dimension of a $D$ dimensional non-linear marked Hawkes process can be expressed as:
	
	\begin{equation}
		\lambda_d(t,m)=\Psi_d \left({\mu_d(t) + \sum_{j=1}^{D} \sum_{\{\forall k |(t_k^j<t)\}} \phi_{dj}(t-t_k^j,m_k^j)}\right)f_d(m),
		\label{Eq:multidimMarkedHawkesGen}
	\end{equation}
	
where $\Psi_d: \R \rightarrow \R^+,$ is a non-negative non-linear function, which is required to be Lipschitz continuous (\cite{bremaud1996stability}), $\mu_d(t):\R^+ \rightarrow \R^+,$ is the exogenous base intensity for the $d$-th dimension, and  $\phi_{dj}(t-t_k^j,m_k^{j}),$  $1\leq d,j\leq D,$ are the kernels that quantify the magnitude of excitation or decay of the intensity for the $d$-th dimension at time $t$ due to the past arrivals $t_k^j,$ and past marks $m_k^{j},$ $\{\forall k | t_k^j < t\}$ in the $j$th dimension. These kernel functions have their support in $\mathbb{R}.$

The intensity function provided in Equation \ref{Eq:multidimMarkedHawkesGen} serves as a comprehensive representation of the marked Hawkes process, wherein the Hawkes kernel $\phi_{dj}$ is influenced by both the histories of marks and event times. However, alternative modelling approaches exist for capturing the impact of marks on the process.

A commonly employed marked intensity function in the literature is expressed as follows:

\begin{equation}
\lambda_d(t,m) = \Psi_d \left({\mu_d(t) + \sum_{j=1}^{D} \sum_{{\forall k |(t_k^j<t)}} \psi_{dj}(m^k_j) \phi_{dj}(t-t_k^j)}\right)f_d(m),
\label{Eq:multidimMarkedHawkes1}
\end{equation}

Similar to Equation \ref{Eq:multidimMarkedHawkesGen}, this formulation involves the conditional ground intensity, which is a function of past event times and corresponding mark values. However, unlike Equation \ref{Eq:multidimMarkedHawkesGen}, it assumes that the excitation kernel can be decomposed into a function of marks and a function of past event time stamps.

In cases where marks are independent and do not influence the evolution of events, the conditional intensity for the marked version is written as:

\begin{equation}
\lambda_d(t,m) = \Psi_d \left({\mu_d(t) + \sum_{j=1}^{D} \sum_{{\forall k |(t_k^j<t)}} \phi_{dj}(t-t_k^j)}\right)f_d(m).
\label{Eq:multidimMarkedHawkes2}
\end{equation}

Here, the marks contribute nothing to the Hawkes kernel, and consequently, they do not impact the ground intensity function, $\lambda_d^g(t)$. Equation \ref{Eq:multidimMarkedHawkesGen}, \ref{Eq:multidimMarkedHawkes1}, and \ref{Eq:multidimMarkedHawkes2} represent different definitions of marked Hawkes processes in terms of mark dependencies. From a model-building perspective, Equations \ref{Eq:multidimMarkedHawkes1} and \ref{Eq:multidimMarkedHawkes2} can be generalized as Equation \ref{Eq:multidimMarkedHawkesGen}. Therefore, it is necessary to consider only Equation \ref{Eq:multidimMarkedHawkesGen} for estimation. To the best of our knowledge, the proposed model is the first non-parametric approach to model the process defined by Equation \ref{Eq:multidimMarkedHawkesGen}.

 Estimating a marked Hawkes process requires obtaining the base intensity function $\mu_d$, and its kernel functions, $\phi_{dj},$ either by assuming a parametric or non-parametric form for the kernels. Similary for estimating the mark density function, $f_d(m)$, one can adopt either a parametric or non-parametric approach. The prevalent method for estimating a Hawkes process typically entails maximizing the log-likelihood function of the process.

\paragraph{The Log-Likelihood Function of Marked Hawkes Process}
	
The categorization of methods used for estimating marked Hawkes processes can be broadly classified into two groups:
\begin{itemize}
\item \emph{Intensity-based approach:} The most commonly used approach for estimating a marked Hawkes process involves deriving the parameters using the conditional intensity function, $\lambda_d(t,m)$. The maximum likelihood estimation (MLE) is the prevailing method in this regard. Notably, this MLE approach has been employed by various researchers, including \cite{du2016recurrent}, \cite{mei2017neural}, and \cite{zuo2020transformer}, among others. However, it's worth noting that MLE methods can be computationally expensive, with their time complexity increasing quadratically with the number of event arrivals. This can pose challenges when dealing with large datasets or complex models. To address this computational cost, alternative approaches like the Expectation-Maximization (EM) method have been used by researchers such as \cite{fox2016spatially} and \cite{yuan2019multivariate}. EM also utilizes the likelihood function as the loss function, but it offers more computationally efficient optimization strategies compared to traditional MLE.

\item \emph{Intensity free approach:}  The use of intensity free approach for estimating marked Hawkes is very scarce. Among them, \cite{xiao2017wasserstein} and \cite{shchur2019intensity}  employ the Wasserstein distance and conditional distribution function, respectively, to estimate the Marked Hawkes process. While \cite{bacry2016first} and \cite{fabre2024neural} use the second-order characteristics equation of a marked Hawkes kernel to estimate the base intensity and kernel of a marked Hawkes process.
	\end{itemize}	
	
In this paper, we use the maximum log-likelihood(MLE) approach to estimate the base intensity intensity and the optimal kernels of the intensity function.

In a  $D$-dimensional non-linear marked Hawkes process, $\{N_d[0,T) \times \mathcal{M}\}$ with the parameters $\boldsymbol{\mu}=[\mu_d(t)]_{D \times 1}$, $\boldsymbol{\Phi}=[\phi_{dj}(t,m)]_{D \times D}$ and $\boldsymbol{f}=[f_d(m)]_{D}$ ; $1 \le d,j \le D$ be the base intensity intensity, kernel function, and mark density function respectively. Let $\boldsymbol{\theta}=[\theta_1,\ldots\theta_l \ldots \theta_L]$ be the set of parameters used to model $\boldsymbol{\mu}$, $\boldsymbol{\Phi}$ and $\boldsymbol{f}$.  These parameters can be estimated by maximizing the log-likelihood function over the events, $\boldsymbol{\mathcal{H}}$ sampled from the process. The log-likelihood (LL) function, $\mathcal{L}$ corresponding to $\boldsymbol{\mathcal{H}}$ for the non-linear marked Hawkes process is given by (see for instance \cite{daley2007introduction}),
\begin{equation}
		\mathcal{L}(\boldsymbol{\theta},\boldsymbol{\mathcal{H}}) = \sum_{d=1}^{D} \left[ \sum_{\{(t^d_n,m^d_n)\}\in \boldsymbol{\mathcal{H}}} \log(\lambda_d(t^d_n,m^d_n)) - \int_0^T \int_{\mathcal{M}} \lambda_d(s,m)ds dm  \right],
	\end{equation}
	
applying the $\lambda_d(t,m)$ from the Equation \ref{Eq:PPIntensityDen} in to the equation,  
\begin{equation*}
\mathcal{L}(\boldsymbol{\theta},\boldsymbol{\mathcal{H}}) = \sum_{d=1}^{D} \left[ \sum_{ \{(t^d_n,m^d_n)\}\in \boldsymbol{\mathcal{H}}} \left\{\log(\lambda^{g}_d(t^d_n)) +\log(f_d(m^d_n)) \right\}- \int_0^T \int_{\mathcal{M}}  \lambda^{g}_d(s) f_d(m)ds dm \right].
\end{equation*}
As the sum of mark density function across the marked space $\mathcal{M}$ is unity, i.e. $ \int_{\mathcal{M}} f_d(m)dm=1$, the above equation can be written as,

\begin{equation*}
		\mathcal{L}(\boldsymbol{\theta},\boldsymbol{\mathcal{H}}) = \sum_{d=1}^{D} \left[ \sum_{\{(t^d_n,m^d_n)\}\in \boldsymbol{\mathcal{H}}} log(\lambda^{g}_d(t^d_n)) - \int_0^T  \lambda^{g}_d(s) ds + \sum_{\{ m^d_n\}\in \boldsymbol{\mathcal{H}}} f_d(m^d_n)\right].
	\end{equation*}
 The provided LL function $\mathcal{L}$ can be expressed as a sum of two terms, $\mathcal{L}= \mathcal{L}_{\lambda} + \mathcal{L}_f $, where $\mathcal{L}_{\lambda}$ is the log-likelihood function for the ground intensity and $\mathcal{L}_f$ is the log-likelihood function for the marks density function.
	\begin{eqnarray}
    \label{Eq:LLLambda}
		\mathcal{L}_{\lambda}(\boldsymbol{\theta_{\lambda}},\boldsymbol{\mathcal{H}}) &=& \sum_{d=1}^{D} \left[ \sum_{\{(t^d_n,m^d_n)\}\in \boldsymbol{\mathcal{H}}} \log(\lambda^{g}_d(t^d_n)) - \int_0^T  \lambda^{g}_d(s)ds  \right], \\
    \label{Eq:LLFunc}
		\mathcal{L}_{f}(\boldsymbol{\theta_{f}},\boldsymbol{\mathcal{H}})&=&\sum_{d=1}^{D}  \left[\sum_{\{m^d_n\}\in \mathcal{S}} \log(f_d(m_n^d)) \right],
	\end{eqnarray}

Where $(\boldsymbol{\theta_{\lambda}},\boldsymbol{\theta_{f}}) \in \boldsymbol{\theta}$ and,  $\boldsymbol{\theta_{\lambda}}$ and $\boldsymbol{\theta_{f}}$  are the parameter sets for the intensity function and density function, respectively. There are no parameters that appear in both $\mathcal{L}_{\lambda}$ and $\mathcal{L}_f$. Thus, both terms can be maximized separately.

From Equation \ref{Eq:multidimMarkedHawkesGen}, the $\lambda_d^g(t)$ in $\mathcal{L}_{\lambda}$ can be written as a combination of a base intensity rate $\mu_d(t)$ and kernel function $\phi_{dj}(t,m)$. Even for the most common parametric form of the $\phi_{dj}(t,m)$, the $\mathcal{L}_{\lambda}$ function may or may not be convex. For example, take the simplest case of the exponential kernel with multiplicative marks, $\phi_{dj}(t,m) = me^{-\beta t}$, $\mathcal{L}_{\lambda}$ won't be a convex function. Hence, the Stochastic Gradient Descent (SGD) method is employed to find the local optimum of the $\mathcal{L}_{\lambda}$ function in the parametric space. To estimate the parameters $\boldsymbol{\theta}$ using SGD, an unbiased estimator of the gradient of $\mathcal{L}_{\lambda}$ with respect to each parameter $\theta_p \in \boldsymbol{\theta_{\lambda}}$ is required and is given by,
	
\begin{eqnarray}\nonumber
		\nabla_{\theta_p} \left(\mathcal{L}_{\lambda}(\boldsymbol{\theta_{\lambda}},\boldsymbol{\mathcal{H}}) \right)
		&=&\sum_{d=1}^{D} \left[ \sum_{\{(t^d_n,m^d_n)\}\in \boldsymbol{\mathcal{H}}} \nabla_{\theta_p} \left(\left\{\log(\lambda^{g}_d(t^d_n))- \int_{t^d_{n-1}}^{t^d_n} \lambda^{g}_d(s) ds  \right\}\right)\right].\label{Eq:LLUnbiased}
	\end{eqnarray}
	
Following the above equation,  for each event points $(t_n^d,m_n^d)$, an unbiased estimator of the gradient of $\mathcal{L}_{\lambda}$ with respect to each of the parameters $\theta_p$  will be,
	
\begin{equation}
		\label{Eq:LLGrad}
		\nabla_{\theta_p}\widehat{\mathcal{L}}_{\lambda}(\boldsymbol{\theta_{\lambda}},t^d_n,m^d_n):=\nabla_{\theta_p}\left(\log(\lambda^{g}_d(t^d_n))- \int_{t^d_{n-1}}^{t^d_n} \lambda^{g}_{d}(s) ds \right),
\end{equation}
	 
where $t^d_n$ and $m^d_n$  are sampled from $\boldsymbol{\mathcal{H}}.$ In this context, we aim to maximize $\widehat{\mathcal{L}}_{\lambda}$ by adopting a non-parametric approach for the kernel function and base intensities. The detailed model is elaborated in Section \ref{Sec:model}.
 
    	\begin{figure}
	    \centering
	    \includegraphics[scale=0.15]{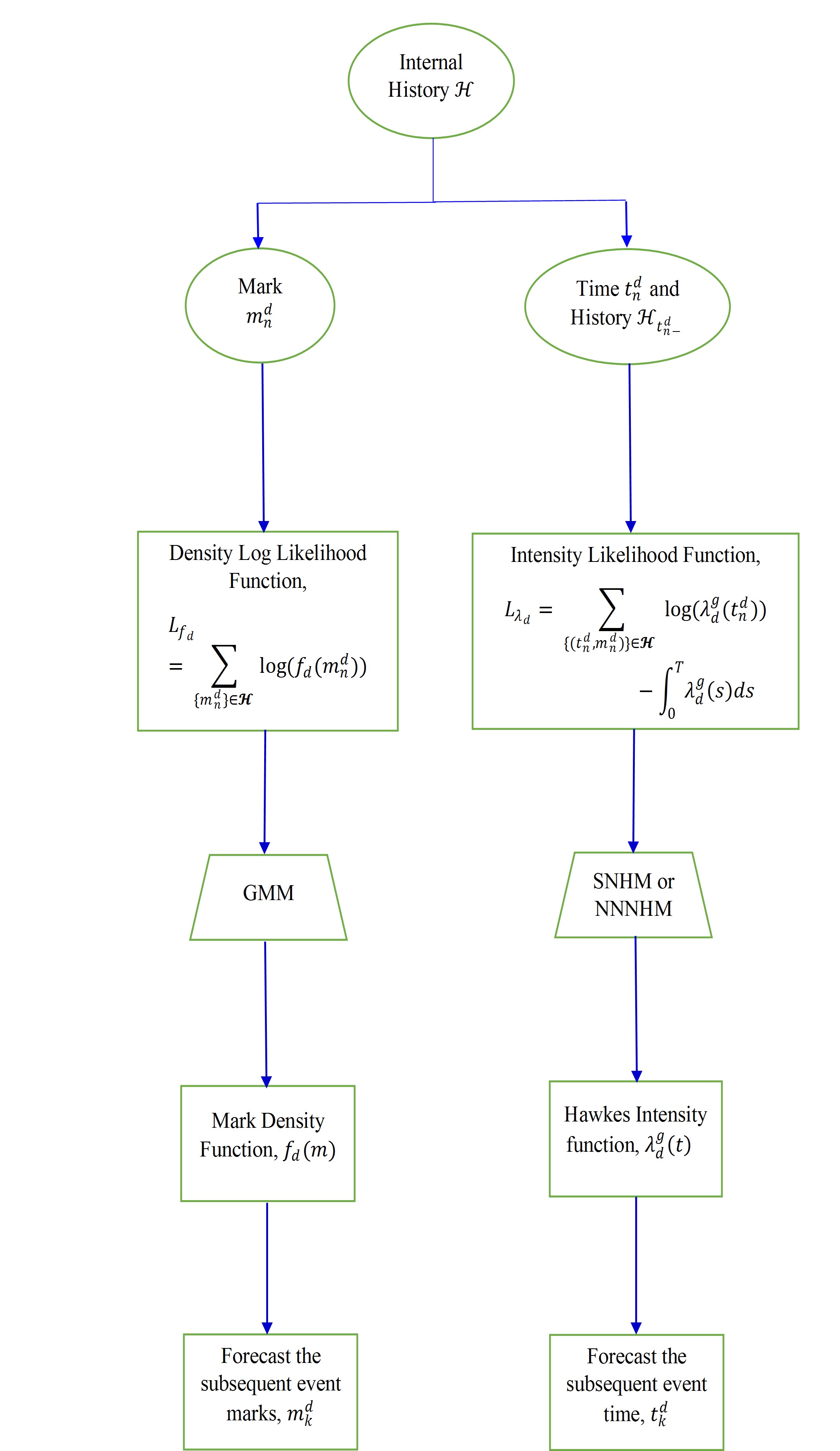}
	    \caption{Flowchart showing process for estimation of Marked Hawkes process}
	    \label{fig:flowchart}
	   \end{figure}

Because the log-likelihood is separable, we can independently fit the conditional mark density. In our approach, we employ the Gaussian mixture model (GMM) as a non-parametric method to estimate the mark density. This entails identifying the GMM parameters, denoted as $\boldsymbol{\theta_{f}},$ that optimize the log-likelihood function $\mathcal{L}_f.$ Further details on this process can be found in Section \ref{Subsec:GMM}.
By combining the non-parametric estimation of the kernel function and base intensities along with the GMM-based approach for $\mathcal{L}_f$, we aim to comprehensively and accurately estimate the parameters for the marked Hawkes process model. Figure \ref{fig:flowchart} provides an overview of the marked Hawkes process estimation for each of the $d$th dimensions. On the left side of the chart, marks $\{m_n^d\} \in \boldsymbol{\mathcal{H}}$  serve as input to the mark density likelihood function, $\mathcal{L}_{f_d}$ from Equation \ref{Eq:LLFunc}, which is then processed by the GMM method to estimate the mark density function, $f_d$. This estimation results in the prediction of subsequent marks for each dimension $d$. On the right side of the chart, both marks and historical time stamps $\mathcal{H}_{t^d_n-}$ are utilized as inputs for the intensity likelihood function, $\mathcal{L}_{\lambda_d}$ given in Equation \ref{Eq:LLLambda}. The SNH with marks, or the NNNH with marks estimation approaches are employed to estimate the network parameters of the ground intensity function, $\lambda^g_d$.  The obtained functional form of $\lambda^g_d$ is used to generate future event times,$t^d_n$.

 \section{Proposed Model}\label{Sec:model}
	
In the context of the internal history $\boldsymbol{\mathcal{H}}$, we estimate the Hawkes conditional intensity function $\lambda_d(t)$ by separately modelling the base intensity $\mu_d(t)$ and the kernel $\phi_{dj}(t,m)$. Estimating the Hawkes kernel is crucial in understanding the underlying temporal and mark dependency with the conditional intensity (as explained in Section \ref{Sec:intro}).
Our approach employs a two-layered neural network with a single hidden layer for approximating the Hawkes kernel $\phi_{dj}$. The choice of using a two-layered neural network is motivated by the universal approximation theorem established by \cite{hornik1989multilayer}, which proves that a multilayered neural network with at least one single hidden layer within a certain group of activation functions can approximate any continuous function in a compact set arbitrarily well. This theorem has been extended to include a wider range of activation functions, including exponential and Rectified Linear Units (ReLu) by \cite{leshno1993multilayer}. 
We refrain from employing a deeper network because estimating the network parameters through maximizing the log-likelihood involves evaluating the time integral of the kernels. Using a shallow network allows for efficient computation of this integral using analytical expressions. In contrast, with deeper networks, the computation of the integral becomes computationally intensive.

This paper proposes two distinct models based on neural networks for estimating the marked Hawkes process --Shallow Neural Hawkes with marks (SNH with Marks) and Neural Network for Non-linear Hawkes with marks (NNNH with marks). The proposed models are the extensions of the SNH (Shallow Neural Hawkes) model \cite{joseph2022shallow} and the NNNH (Neural Network for  Non-linear Hawkes) model \cite{joseph2023neural}. While both the SNH and NNNH use only event times for estimating the Hawkes kernel, the extended marks model incorporates both event times and their corresponding marks, making it a more versatile estimation of the kernel function.

 \paragraph{Shallow Neural Hawkes with Marks} 
In their research, \cite{joseph2022shallow} proposed the Shallow Neural Hawkes (SNH) method, which is a neural network-based approach designed for estimating the kernels of a linear Hawkes process, specifically for process with excitation or positive kernels. 
Expanding upon the SNH model, the \textit{Shallow Neural Hawkes with marks}(SNH with marks) incorporates both event times and their corresponding marks in the Hawkes intensity function. This enhancement enables the method to handle Hawkes processes involving temporal information (event times) and mark information.

The SNH with marks is specifically designed for the linear Hawkes process, which makes it suitable only for positive or excitatory kernels, constraining the kernel function $\hat{\phi}_{dj}(t,m)$ to be bounded in $\R^+$. A Linear Hawkes process implies linearity in intensity function, $\lambda^g_d(t)$. Therefore, in the expression of  $\lambda^g_d(t)$ in Equation\ref{Eq:multidimMarkedHawkesGen}, $\Psi_d$ is an identity function. Thus, the  estimated ground intensity function $\hat{\lambda}^{g}_d(t)$ can be expressed as follows: 
    
    \begin{equation*}
        	\hat{\lambda}^{g}_d(t)=\hat{\mu}_d(t) + \sum_{j=1}^{D} \sum_{\{\forall k |(t_k^j<t)\}} \hat{\phi}_{dj}(t-t_k^j,m^{j}_k),    
    \end{equation*}
 
 where each $\hat{\mu}_d(t),$ and $\hat{\phi}_{dj}(t-t_k^j,m^{j}_k), \, d=1,\ldots,D,$ are approximated using a neural network. Given $\boldsymbol{\mathcal{H}}$, the kernel $\hat{\phi}_{dj}(t,m):\R^+ \times \R \rightarrow \R^+$ is modelled as,
	\begin{eqnarray}\label{Eq:NNModel}
		\hat{\phi}_{dj}(t,m) &=& \varPi \circ A_2 \circ \varphi \circ A_1,
	\end{eqnarray}
	where $A_1$ is the hidden layer, $A_1:\R^+ \times \R \rightarrow \R^P$ and $A_2$ is the output layer, $A_2:\R^P \rightarrow \R^+$. More precisely,
	\begin{eqnarray*}
		A_1\left(\boldsymbol{x}\right) &=& W_1\boldsymbol{x} +b_1 \text{ for } \boldsymbol{x} \in \R^{2 \times 1}, W_1 \in \R^{P \times 2} \text{ and } b_1 \in \R^P,\\
		A_2(y) &=& W_2y+b_2 \text{ for } y \in \R^{P}, W_2 \in \R^{1 \times P} \text{ and } b_2 \in \R,
	\end{eqnarray*}
	where $P$ is the number of neurons. Also, $\varphi:\R \rightarrow \R^+,  1 \le i \le P$ is the ReLU activation function given by,
	\begin{equation*}
		\varphi(y^i) = \max(y^i,0),
	\end{equation*}
	and function $\varPi:\R \rightarrow \R^+$, is the \textit{exponential} function. We take $W_1=\begin{bmatrix} a^1_1 & a^2_1 & \ldots &a^P_1 \\ a^1_2 & a^2_2 & \ldots &a^P_2  \end{bmatrix}^T$, $W_2=\left[ a^1_3,a^2_3,\ldots,a^P_3 \right]$, and $b_1=\left[ b^1_1,b^2_1,\ldots,b^P_1 \right]$. Thus, the kernel function $\hat{\phi}_dj$ can be written as,
	\begin{equation}\label{Eq:SNHmodel}
		\hat{\phi}_{dj}(t,m) = \text{exp}\left(b_2 + \sum_{i=1}^{p}a^i_3\text{max}\left( a^i_1 t + a^i_2 m +b^i_1, 0\right)\right).
	\end{equation}

 \paragraph{Neural Network for Non-Linear Hawkes with Marks} 
To overcome the limitation of the SNH with marks, which is confined to linear Hawkes processes with only excitation kernel functions, we propose a novel model called \textit{Neural Network for Non-linear Hawkes with Marks} (NNNH with marks). Adapted from the Neural Network for Non-linear Hawkes (NNNH) method introduced by \cite{joseph2023neural}, the NNNH with marks can model both excitatory and inhibitory kernels in a non-linear marked Hawkes process. This means that the kernel function $\hat{\phi}_{dj}$ can take negative values with the range of the function mapping to $\R$. 
To guarantee that the intensity function $\lambda^{g}_d(t)$ remains positive, the function $\Psi_d$ in Equation \ref{Eq:multidimMarkedHawkesGen} is selected as the ReLU function. By employing the RELU function, the estimated Hawkes intensity $\hat{\lambda}^{g}_d(t)$ is guaranteed to remain positive throughout the estimation, thus adhering to the necessary constraint. The framework is flexible enough to consider the base intensity $\mu_d(t)$ as a function of time. Thus the estimated Hawkes ground intensity function $\hat{\lambda}^{g}_d(t)$ for the NNNH with marks is given by, 
    \begin{equation}\label{Eq:NonLinearHawkes}
		\hat{\lambda}_d^{g}(t)= \max \left({\hat{\mu}_d(t) + \sum_{j=1}^{D} \sum_{\{\forall k |(t_k^j<t)\}} \hat{\phi}_{dj}(t-t_k^j,m^{j}_k)}, 0\right),
	\end{equation} 

where each $\hat{\mu}_d(t),$ and $\hat{\phi}_{dj}(t-t_k^j,m^{j}_k), \, d=1,\ldots,D,$ are approximated using a neural network. The network utilized to represent $\hat{\phi}_{dj}(t,m)$ shares similarities with the one outlined in Equation \ref{Eq:SNHmodel}. However, in the case of the NNNH network, the function $\Pi$ takes the form of an identity function instead of an exponential one. This selection guarantees that $\hat{\phi}_{dj}$ effectively maps from $\mathbb{R}^+ \times \mathbb{R}$ to $\mathbb{R}$.
With this modification in $\hat{\phi}_{dj}$, the NNNH with mark can effectively accommodate excitation and inhibition kernels, making it more versatile for modelling a broader range of processes involving diverse dependencies between past arrivals and their corresponding marks. The kernel function $\hat{\phi}_{dj}(t,m)$ given $\boldsymbol{\mathcal{H}}$ is modelled as,
 \begin{eqnarray}\label{Eq:NNNHmodel}
     		\hat{\phi}_{dj}(t,m) = b_2 + \sum_{i=1}^{p}a^i_3\text{max}\left( a^i_1 t + a^i_2 m +b^i_1, 0\right).
 \end{eqnarray}

  \begin{remark}\label{rem:Comparison}
This paper employs two methodologies to estimate the marked Hawkes process: SNH with marks for linear marked Hawkes and NNNH with marks for both linear and non-linear marked Hawkes. Despite NNNH with marks being capable of estimating the linear marked Hawkes process, SNH with marks is preferred due to its lower computational cost. When there is ambiguity in determining whether an event results in excitation or inhibition of the ground intensity in a dataset $\boldsymbol{\mathcal{H}}$, it is recommended to use NNNH with marks. However, if it is certain that the arrival of an event leads to the excitation of the ground intensity process, then SNH with marks is the preferable choice. 
  \end{remark}  
    
    The neural network used for both estimation models consists of a hidden layer of $P$ neurons and a single output layer, resulting in a parameter dimension of $4P+1$ for a one-dimensional marked Hawkes setup. For a $D$-dimensional marked Hawkes process with a constant base intensity, the number of parameters can go up to $(4P+1)(D^2)$. If we also consider the case of non-constant base intensity, $\mu_d(t)$, then the number of parameters to be estimated will be  $(4P+1)(D^2+D).$

    To estimate the parameter set $\boldsymbol{\mathcal{\theta}_{\lambda}}$ encompassing both the kernels and the base intensity parameters ($(4P+1)(D^2+D)$ parameters), we use the ground intensity log-likelihood function $L_{\lambda}$ given in Equation \ref{Eq:LLLambda} across the observed dataset $\boldsymbol{\mathcal{H}}$. The computation of $L_{\lambda}$ using the neural network approximation of $\lambda^g_d(t)$ in both the models necessitates obtaining the integrated intensity function and log intensity function in $L_{\lambda}$, the details of which are given in Appendix \ref{App:ApprLL}.
    
    Post obtaining the log-likelihood function $L_{\lambda}$, in order to find the optimal network parameter set $\boldsymbol{\hat{\theta}_{\lambda}}$, which provides a local maximum for the $L_{\lambda}$, we employ batch stochastic gradient descent (SGD) combined with Adam optimization algorithm introduced by \cite{kingma2014adam}. The gradients of the  $L_{\lambda}$ with respect to each of the parameters in the  $\boldsymbol{\mathcal{\theta}_{\lambda}} $ can be computed using Equation \ref{Eq:LLUnbiased}. These gradients acquired through the batch SGD iteratively update the parameters in $\boldsymbol\theta_{\lambda}$, leading to the identification of the optimal parameter set, $\boldsymbol{\hat{\theta}_{\lambda}}$  maximizing the $L_{\lambda}$, providing the optimal estimations for the kernels and the base intensity.

	\subsection{Gaussian Mixture Model}{\label{Subsec:GMM}}
	Gaussian mixture model (GMM) represents a class of mixture models used to estimate density functions. According to \cite{murphy2012machine} and \cite{mclachlan1988mixture}, mixture models are probabilistic models where a distribution can be represented as the weighted sum of two or more base distributions. GMM specifically uses multivariate Gaussian distribution as the base distribution. GMM with a sufficiently large number of mixture components (Gaussian distributions) can approximate any continuous density function defined on $\R$ (\cite{murphy2012machine}). 
  This paper utilizes a GMM density function to approximate the $f_d(m),$ mark density function in the $d$th dimension. While the marks, represented as $[m^1, \ldots, m^D]$, may exhibit correlations, our specific assumption is that marks within a dimension are independent of those in other dimensions. For instance, the trade volumes for buy and sell market orders are assumed to be independent in the context of trade orders. Consequently, the marks within each dimension are treated as independent and identically distributed (i.i.d.). 
  The estimated GMM density function $\hat{f}_d(m|\boldsymbol{\theta_f}),$ corresponding to the $d$th dimensional marks $m=\{m^d_n\}_{n \ge 1}$ sampled from $\boldsymbol{\mathcal{H}}$, given the parameter set $\boldsymbol{\theta_f}= \left\{(z_j,\mu_j,\sigma^2_j)\right\}$, is expressed as follows:

	\begin{equation}
		\hat{f}_d(m|\boldsymbol{\theta_f}) \approx \sum_{j=1}^{k} z_j F(m|\mu_j,\sigma^2_j), 	
	\end{equation}
	where $F(m|\mu_j,\sigma_j^2)$ is the density function of the $j$th Gaussian distribution given by,
 $$F(m|\mu_j,\sigma_j^2) = \frac{1}{\sqrt{2\pi\sigma_j^2}}\exp{-\left(\frac{(m-\mu_j)^2}{2\sigma_j^2}\right)}.$$
 with parameters $(\mu_j,\sigma_j^2)$, $k$ is the number of Guassian distributions and $z_j, 0 \le z_j \le 1$ is the weight parameter. The weight parameter, $z_j$ should satisfy $\sum_{j=1}^k z_j =1$. The parameters of the GMM, $\boldsymbol{\theta_f}$ are estimated by the expectation maximization (EM) algorithm explained by \cite{dempster1977maximum}.

\section{Experiments and Results} \label{Sec:results}
	
	The evaluation process commences with a comprehensive analysis of synthetic datasets, encompassing both one-dimensional and multi-dimensional Hawkes processes with continuous marks.The effectiveness of our neural network-based approaches, SNH with marks and NNNH with marks is scrutinized against the vanilla models of SNH and NNNH\footnote{The code and the data that has been used in this section is given in https://github.com/sobin-joseph/Marked-Hawkes-Estimation}.

    The evaluation extends to practical applications, specifically utilizing our neural network approach on cryptocurrency trading data. This data includes detailed tick-by-tick sell and buy market orders for Bitcoin US Dollars (BTC-USD) and Ethereum US Dollar (ETH-USD) pairs. The neural network-based approach is applied to unveil the causal relationships concerning arrival times and marks within the specified pairs.
    
    To ensure the robustness of our methodology, we initiate the evaluation with data preprocessing steps before directly implementing the estimation technique. Moreover, we adopt a scaling mechanism for the dataset, which will be elaborated upon in the subsequent sections. The choice of initial parameters and hyperparameter selection are also addressed, including factors such as the number of hidden neurons, learning rates, and the stopping criteria. These aspects collectively contribute to appropriately evaluating our neural network-based approach's performance and its practical applicability in different scenarios.
    
    \subsection{Simulation of Marked Hawkes Process}
     To generate a synthetic dataset for a multivariate marked Hawkes process, it is crucial to define a simulation algorithm for the process. For simulating a Hawkes process, the most widely used algorithm is \textit{Ogata's modified thinning algorithm} proposed by \cite{ogata1981lewis}. Ogata's modified thinning algorithm builds upon the thinning algorithm initially introduced by \cite{lewis1979simulation} to simulate the point process. Thinning algorithms rely on the conditional intensity function to simulate successive intervals. For the simulation of a marked point process \cite{daley2007introduction} extended the thinning algorithm of \cite{lewis1979simulation} to incorporate the marks. The resulting thinning algorithm for the marked Hawkes process given in Algorithm \ref{alg:ThinningMarked}, is employed to simulate the marked Hawkes process. Similar to the aforementioned thinning algorithms, this algorithm for the marked point process follows a sequential approach. It utilizes the ground conditional intensity,$\lambda^g_d(t)$ to select the next time point. Subsequently, after selecting a new event time, the corresponding mark is chosen using the density function of the marks,$f_d(m)$.
     	 \begin{algorithm}[!ht]
			\SetAlgoLined
			
			\textbf{Input}: base intensity$\{\mu_d\}_{d=1}^D$ , kernel function$\{\phi_{dj}\}_{d,j=1}^D$, market density function$\{f_{d}\}_{d=1}^D$, end time $\boldsymbol{T}$\;
            \textbf{Output}: Event times $\boldsymbol{\{\tau_d\}_{d=1}^D}$ and marks $\boldsymbol{\{m_{d}\}_{d=1}^D}$\; 
			Initialize $t \gets 0$, $\boldsymbol{\{\tau_d\}_{d=1}^D} \gets \emptyset$, $\boldsymbol{\{m_{d}\}_{d=1}^D} \gets \emptyset$\;
			\While {$t<\boldsymbol{T}$}{
                Initialize $M(t)\gets 0$ \;
				\For{ d= 1 to D}{
					$\lambda^g_d(t) \gets \max\left(\mu_d, \mu_d + \displaystyle\sum_{j=1}^{d} \displaystyle\sum_{t\le \boldsymbol{\tau_j}} \phi_{dj} (t-\boldsymbol{\tau_j},\boldsymbol{m_{j}})\right)$\;
                    $M(t) \gets M(t) + \lambda^g_d(t)$\;
                    }
				$D \sim \mathbf{U}(0,1)$\;
				$s \gets t-\frac{\log(D)}{M(t)}$\;
                $U \sim \mathbf{U}(0,1)$\;
                Initialize $\lambda^g(s) \gets 0$ \;
				\For{ d= 1 to D}{
                    $\lambda^g_d(s) \gets \max\left(0, \mu_d + \displaystyle\sum_{j=1}^{d} \displaystyle\sum_{s< \boldsymbol{\tau_j}} \phi_{dj} (s-\boldsymbol{\tau_j},\boldsymbol{m_{j}})\right)$\;
                    $\lambda^g(s) \gets \lambda^g(s)+ \lambda^g_d(s)$ \;
                    \eIf { $\frac{\lambda^g(s)}{M(t)}< U $ }{
                    $t \gets s$ \;
                    $\boldsymbol{\tau_d} \cup s$ \;
                    $F \sim \mathbf{U}(0,1)$\;
                    $m \gets f_{d}^{-1}(F)$\;
                    $\boldsymbol{m_{d}} \cup m$ \;
                    }
                    
                }
				{pass\;}

				}
			\caption{Thinning algorithm for marked Hawkes process}
            \label{alg:ThinningMarked}
		\end{algorithm}	
	\subsection{Data preprocessing and choice of hyper-parameters:}
    Before conducting the experiments, we partition the dataset $\boldsymbol{\mathcal{H}}$ into training, validation, and test sets following a 70:15:15 ratio. It is important to ensure that the partition preserves the event times' chronological order. For both the validation and test sets, all the historical information leading up to the current event is considered, starting from the inception of the data. Prior to the dataset division, we apply scaling to enhance the data's suitability for analysis. This scaling procedure is aimed at transforming the dataset into a more canonical form, which, in turn, reduces the inherent variability. By reducing variability, we achieve a twofold effect: diminishing the generalization error and decreasing the model's size, which, in our case, pertains to the number of neurons required to effectively train the model (\cite{goodfellow2016deep}).
    
    This preprocessing procedure involves both event times and their corresponding marks. Given dataset  $\boldsymbol{\mathcal{H}}\, = \,\{(t^d_n,m^d_n)\}_{n \ge 1},$ where $t^d_n\in[0,T)$ and $m^d_n\in \mathcal{M}$ denote the ordered arrivals and their corresponding marks for the $d$th dimension ,the following definitions are relevant:

        \[t^d_{\max}=\max\left\{t^d_n\right\}_{n \ge 1},
        T_{\max}=\max\left\{t^d_{\max}\right\}_{d=1}^D, 
         N=\sum_{d=1}^{D}N_d(T_{\max}) \text{ and }
         m^{d}_{\text{mean}}= \dfrac{\sum\limits_{m^d_n \in \boldsymbol{\mathcal{H}}}m^{d}_n}{N}\]
  
    Consequently, each event time $t^d_n$ and its corresponding mark $m^d_n$ are scaled as follows:
  
        \[\hat{t}^d_n = t^d_n \frac{N}{T_{\max}},\]
         \[\hat{m}^{d}_n = m^{d}_n\frac{1}{m^{d}_{\text{mean}}}.\]
    
    The scaled dataset $\boldsymbol{\mathcal{\hat{H}}}\, = \,\{(\hat{t}^d_n,\hat{m}^d_n)\}_{n \ge 1},$ will be the input to the proposed model.
    
    \paragraph{Initialization of network parameters and choice of batch size and hyperparameters:}
   Proper initialization the parameters of the proposed models is of paramount importance, as it significantly impacts the performance of most algorithms. According to \cite{goodfellow2016deep}, the choice of initial values plays a critical role in determining whether an algorithm converges successfully or not. In certain cases, unstable initial inputs can even lead to the complete failure of the model.

    To address this concern, we adopt distinct initialization strategies for SNH with marks and NNNH with marks. For SNH with marks, the parameters outlined in Equation \ref{Eq:SNHmodel} are initialized as follows:
    
	   $a_1^i \sim \mathbf{U}(-0.5 , 0.5),$
	   $a_2^i \sim \mathbf{U}(-0.2 , 0.2),$
      $a_3^i \sim \mathbf{U}(-1.0 , 0.0),$
      $b_1^i \sim \mathbf{U}(0.0 , 0.03),$ and 
	   $b_2^i \sim \mathbf{U}( -0.1,0),$
 
    where $\mathbf{U}(a , b),$ denotes uniform distribution between $a$ and $b.$ 

    For the NNNH with marks, the parameters given in Equation \ref{Eq:NNNHmodel}  are initialized as:

	   $a_1^i \sim \mathbf{U}(-0.7, 0),$
	   $a_2^i \sim\mathbf{U}(-0.2 , 0.2),$
      $a_3^i \sim \mathbf{U}(0 , 1.0),$ 
      $b_1^i \sim \mathbf{U}(0.0 , 0.25),$ and
      $b_2^i = 0.$

    The initial base intensity for both the models is set to be initialised $\mu = 1$ (For cases of varying base intensity in the NNNH with marks approach, the parameters of base intensity are initialised as explained in \cite{joseph2023neural}). We maintain a fixed hidden layer size of sixty-four neurons in both the SNH with marks and NNNH with marks. This choice of the number of neurons results from a careful balance between computational time requirements and achieving the optimum log-likelihood.

    Our stochastic gradient calculations are performed using a batch size of one hundred. We employ distinct learning rates for updating parameters in the hidden and output layers, as we have observed faster convergence with this configuration. The specific learning rates adopted for our experiments are as follows:

    \begin{itemize}
        \item For the SNH with marks model, we use a learning rate of $2 \times 10^{-2}$ for updating network parameters in the output layer and $2 \times 10^{-3}$ for updating parameters in the hidden layer.
        \item For the NNNH with marks,  $5 \times 10^{-3}$ and $5 \times 10^{-4}$ are used to update the parameters in the output and hidden layers, respectively. 
    \end{itemize}
    In both models, the learning rate for updating $\mu$ is $1 \times 10^{-3}$. These learning rate choices have contributed to efficient and effective convergence during the training process for both the SNH with marks and NNNH with marks.
    \paragraph{Stopping criteria:}
    We employ a stopping criterion to prevent the algorithm from underfitting or overfitting the dataset. For both the SNH with marks and NNNH with marks, we utilize the number of iterations during which the validation log-likelihood did not show improvement as the stopping criterion. In this context, we set the number of iterations for the stopping criteria to a fixed value of $10$. This ensures that the optimization process halts after $10$ consecutive iterations without observable enhancements in the validation log-likelihood. 

	\subsection{Synthetic Data Experiments}
	\label{Subsec:UniEst}
    This section assesses the effectiveness of the proposed models on simulated datasets of a marked Hawkes process. We explore three distinct scenarios: a one-dimensional linear marked Hawkes process with a combined (or coupled) kernel function, a one-dimensional non-linear marked Hawkes process with decoupled kernel function (separable function of time and marks), and a multivariate marked Hawkes with decoupled kernel functions.  

    For generating synthetic datasets for each scenario, we employ the thinning algorithm for marked Hawkes process outline in Algorithm \ref{alg:ThinningMarked}.  
    The evaluation of the proposed approaches involves the following measures: as an initial performance metric, we compare the error between the known theoretical kernel and the kernel estimated using the proposed methods- both visually and quantitatively through an error plot. We analyze the absolute difference between the theoretical kernel and the estimated kernel at all regions in the error plot. Next, we compare the predictions of the estimated marked Hawkes process for arrival times with the vanilla models. This comparison is facilitated using Quantile-Quantile (QQ) plots for the test dataset.

    QQ plots validate the accuracy of interarrival time predictions from the estimation method. When predicting the next arrival time with a 90\% confidence level based on the current history, ideally, 90\% of the observed next arrivals should occur within the predicted time, and at least 10\% should occur beyond it. If all subsequent arrivals fall within the predicted time, the algorithm is overly conservative, predicting too far into the future. If fewer than 90\% of subsequent arrivals align with the predicted horizon, the predictions are too soon. A well-performing prediction model should exhibit Q\% correct predictions at a Q\% confidence level. A correctly fitted model is represented by an ideal QQ plot, which is a 45-degree line running through the origin.
\subsubsection{One-dimensional Marked Hawkes Process}
\begin{figure}[ht]
     \centering
     \begin{subfigure}[b]{0.45\textwidth}
         \centering
         \includegraphics[width=\linewidth]{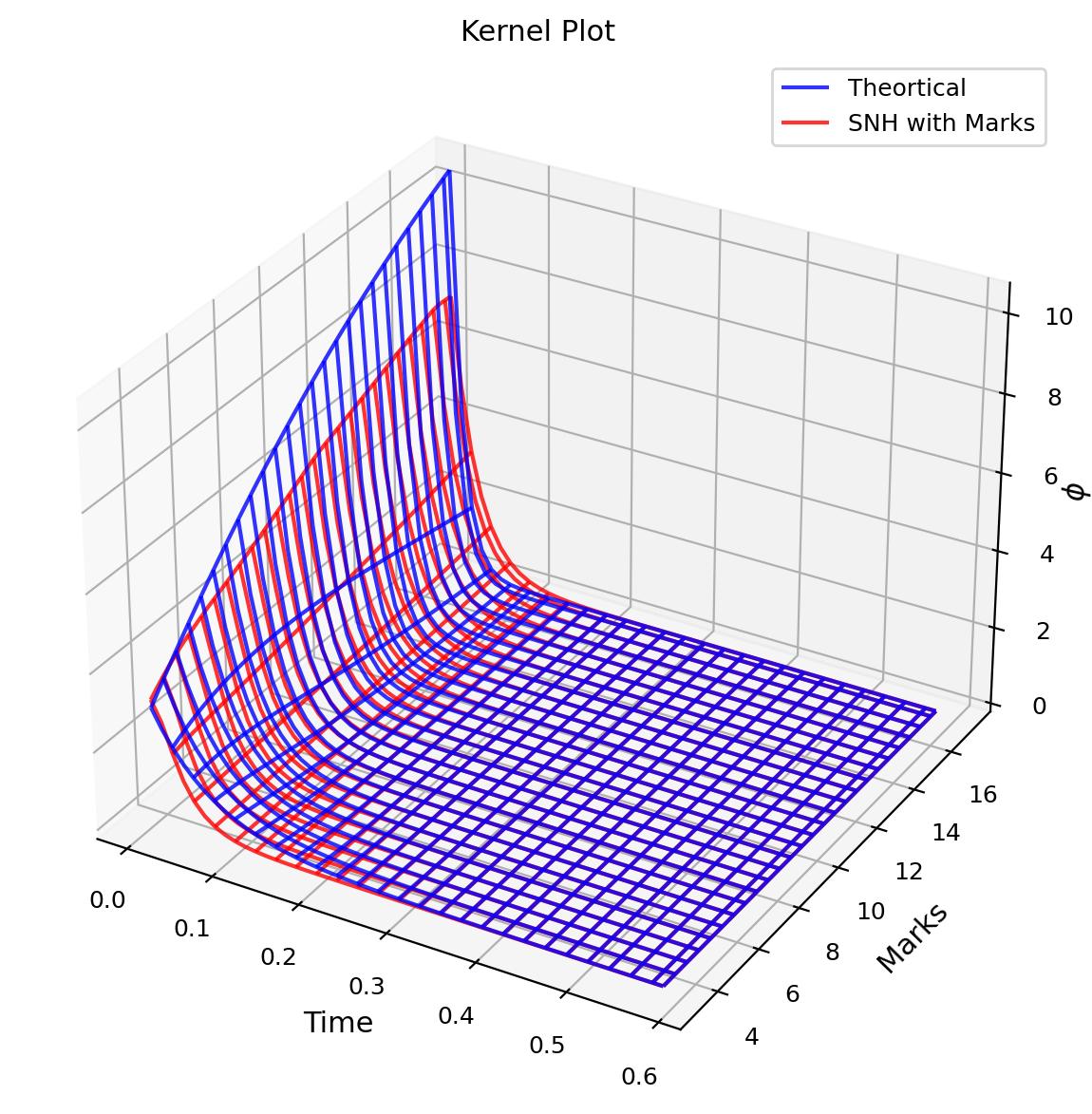}
        \caption{Estimated and theoretical kernel}\label{Fig:real1}
     \end{subfigure}
     \hfill
     \begin{subfigure}[b]{0.45\textwidth}
         \centering
         \includegraphics[width=\linewidth]{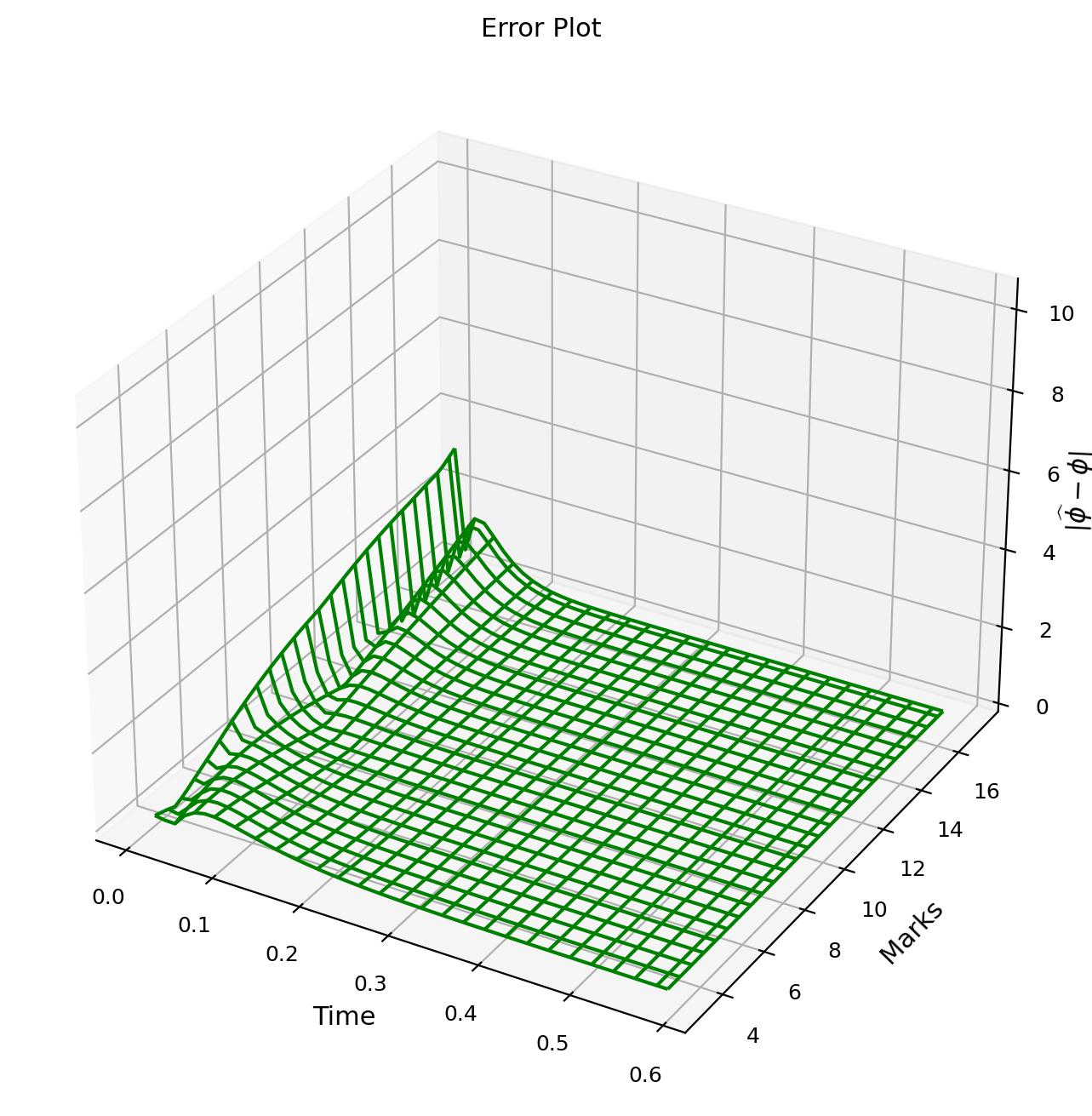}
         \caption{Error plot}\label{Fig:error1}
    \end{subfigure}
    \caption{(a) describes the theoretical and estimated kernel obtained using SNH with marks for a one-dimensional Linear Hawkes process, while (b) presents the absolute error between the theoretical and estimated kernel expressed as $|\hat{\phi} - \phi|$}
    \label{Fig:expkernel}
    \end{figure}

   \paragraph{Marked Hawkes with coupled kernel function:} 
  This example uses the coupled (or combined) kernel function of the generalized marked Hawkes conditional intensity, as described in Equation \ref{Eq:multidimMarkedHawkesGen}. Specifically, we focus on a scenario representing a linear marked Hawkes process. In this context, we investigate situations where $\phi(t,m)$ is not a separable function of time $t$ and marks $m$. To the best of our knowledge, no other non-parametric models can estimate such kernels. The marked Hawkes process is simulated using the following functions: 
    \begin{itemize}
        \item Kernel function, $\phi(t,m) = me^{-t(1 + 5m)},$
        \item Base intensity, $\mu =0.7$, and
        \item Mark distribution, $f(m) = \frac{1}{0.5 m \sqrt{2 \pi}} e^{-\frac{(\log m)^2}{2 \times 0.5^2}}$.
    \end{itemize}
    Using Algorithm \ref{alg:ThinningMarked} with specified parameters, we conducted a simulation for the time interval $(0,2000]$, resulting in 1743 arrivals. The dataset is generated using a linear Hawkes process, and therefore, for estimation, we can consider either SNH with marks or NNNH with marks. SNH with marks is preferred due to its lower computational demand, as explained in Remark \ref{rem:Comparison}. The SNH estimation reaches the stopping criteria in approximately 690 seconds. Figure \ref{Fig:expkernel} displays the theoretical, estimated, and error kernels. Despite using a limited number of data points for fitting, the proposed model effectively estimates the kernel. The error plot in Figure \ref{Fig:expkernel} depicts the absolute difference between theoretical and estimated kernels ($|\phi-\hat{\phi}|$), demonstrating the model's capability to capture excitation in both time and mark space.
    
    \begin{figure}[ht]
        \centering
        \includegraphics[scale=0.3]{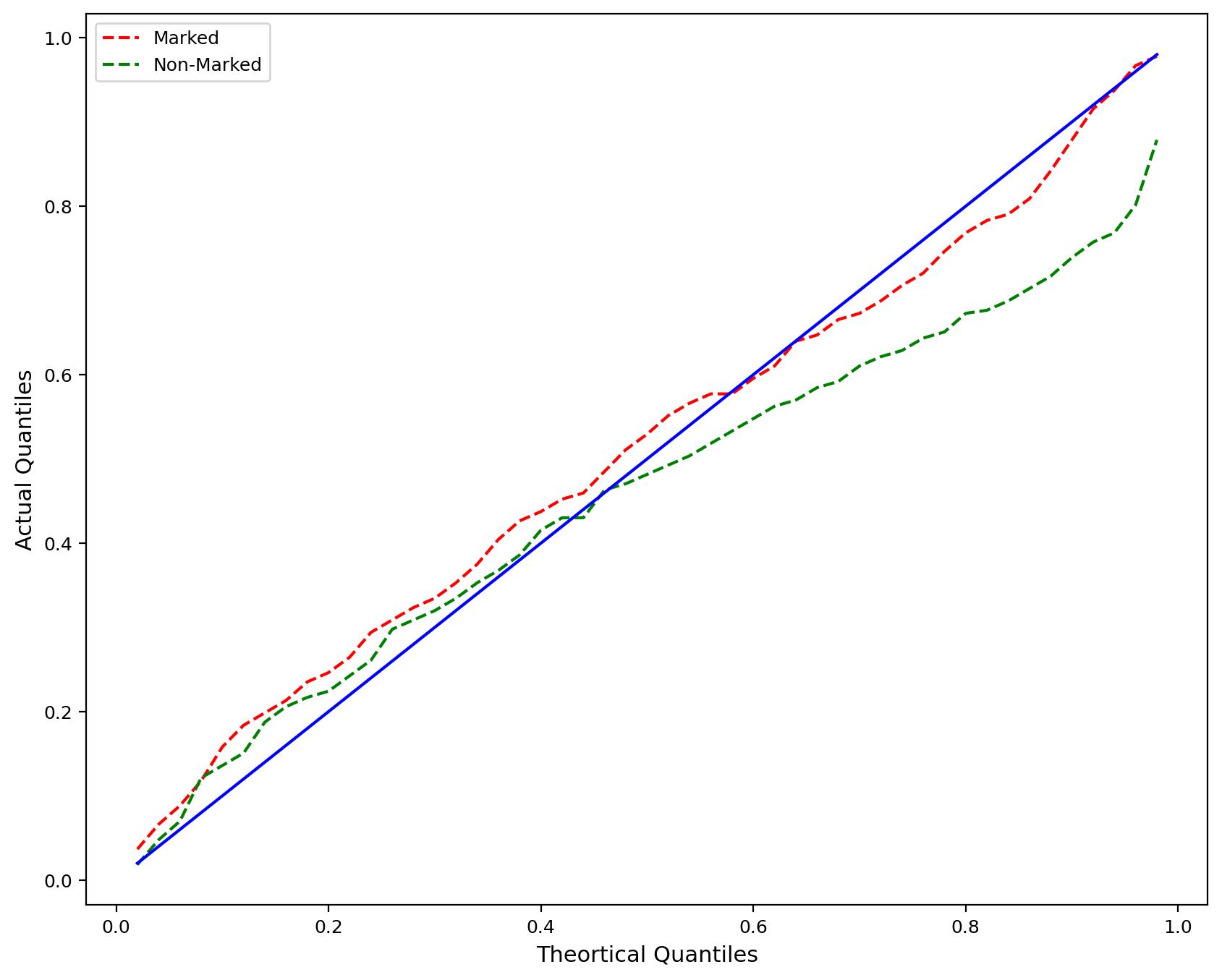}
        \caption{QQ plot providing a comparative analysis between SNH and SNH with marks estimation}
        \label{Fig:SNHqqplot}
    \end{figure}

Figure \ref{Fig:SNHqqplot} presents a QQ plot analysis comparing the performance of SNH with marks and vanilla SNH. The plot indicates that vanilla SNH tends to predict shorter interarrival times more often, whereas SNH with marks accurately captures the distribution of interarrival times. This comparison suggests that the SNH with marks model outperforms the vanilla SNH model in predicting the linear marked Hawkes process.
    
\begin{figure}[ht]
        \centering
        \includegraphics[scale=0.3]{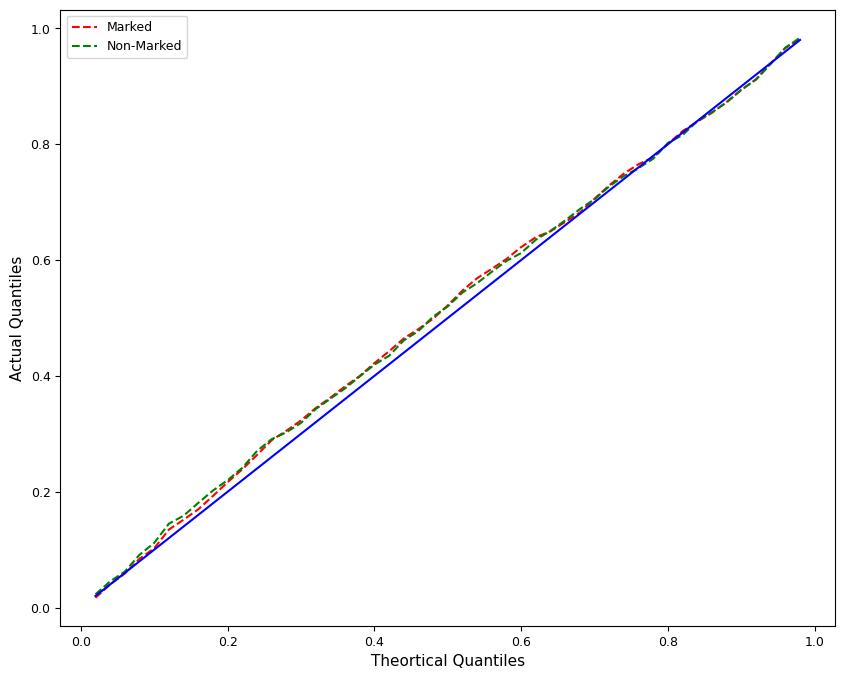}
        \caption{QQ plot comparing the interarrival times obtained from NNNH and the NNNH with marks Estimation.}
        \label{Fig:NNNHqqplot}
    \end{figure}

    \paragraph{Marked Hawkes process with decoupled kernel function:} In our second example, we consider a non-linear marked Hawkes process with decoupled (or separable) kernel function, outlined by the intensity function in Equation \ref{Eq:multidimMarkedHawkes2}. A decoupled kernel function implies that the kernel has distinct functions for its temporal and mark components. For the estimation of non-linear marked Hawkes, it is necessary to use the NNNH with marks, as the SNH with marks is unsuitable for non-linear Hawkes. We use the following parameters for the functions given in the intensity Equation \ref{Eq:multidimMarkedHawkes2}:
    \begin{itemize}
        \item Kernel function, $\phi(t,m) = \psi(m) \phi(t) = \log(m) (-0.4e^{-2t}) ,$
        \item base intensity, $\mu =0.9$, and 
        \item Mark distribution, $f(m) = \frac{1}{m \sqrt{2 \pi}} e^{-\frac{(\log m-0.5)^2}{2 \times 1^2}}$.
    \end{itemize}
    \begin{figure}[ht]
     \centering
     \begin{subfigure}[b]{0.45\textwidth}
        \centering
        \includegraphics[width=\linewidth]{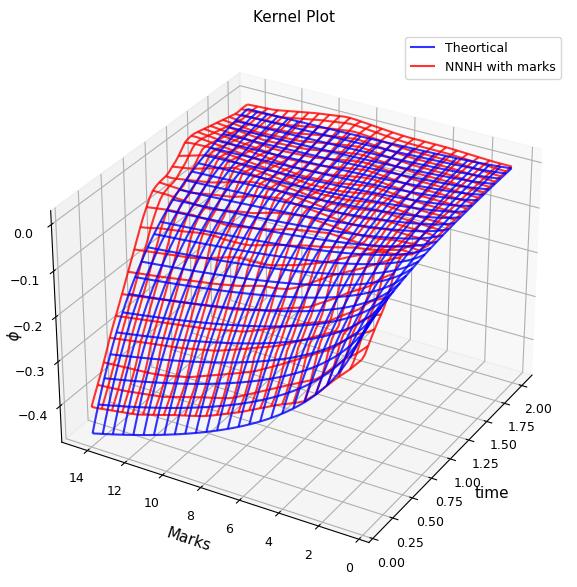}
        \caption{Theoretical and estimated kernel}\label{Fig:real2}
     \end{subfigure}
     \hfill
     \begin{subfigure}[b]{0.45\textwidth}
         \centering
        \includegraphics[width=\linewidth]{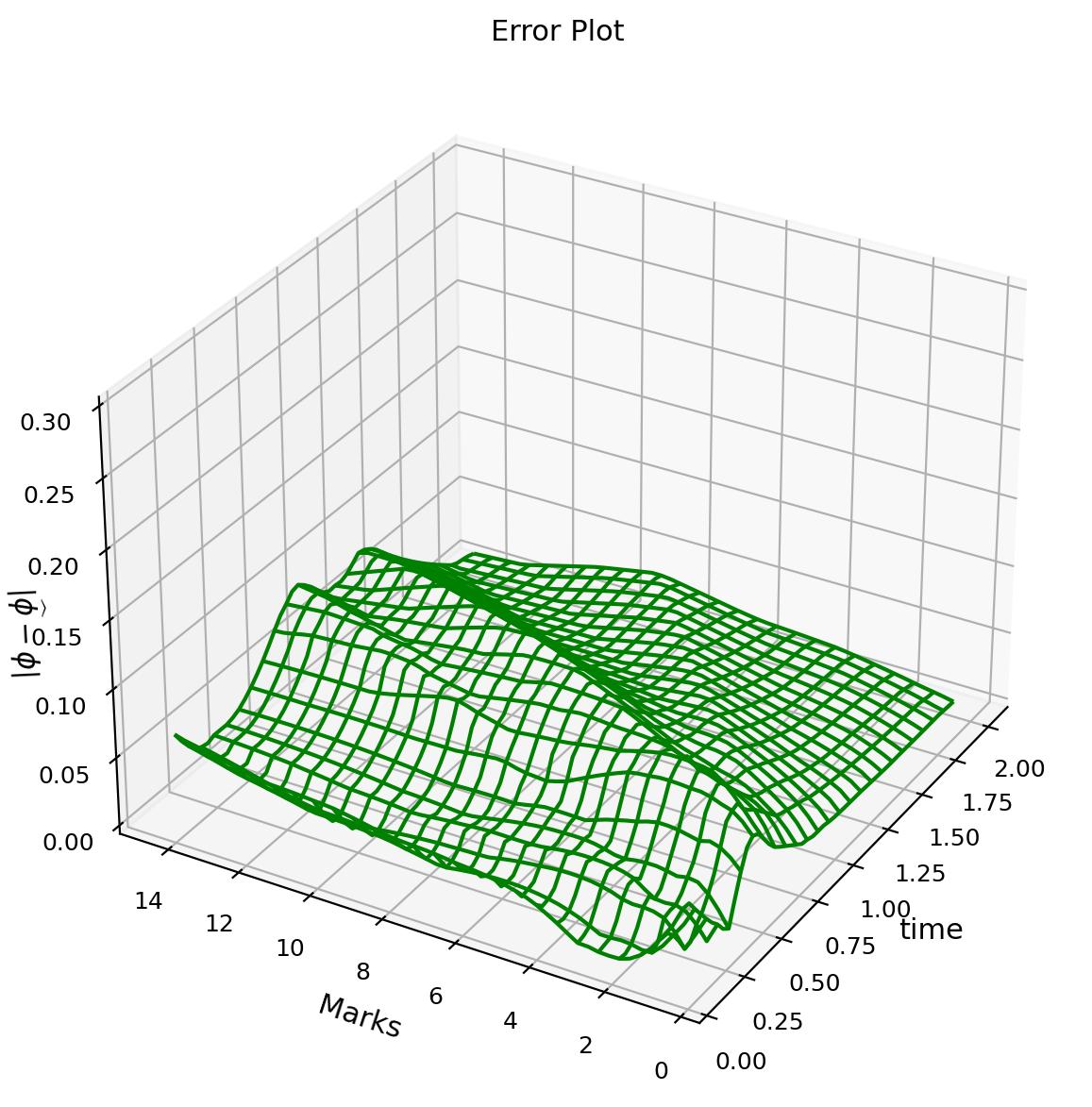}
        \caption{Error plot}\label{Fig:error2}
    \end{subfigure}
    \caption{(a) shows the theoretical kernel and kernel estimated using NNNH with Hawkes for a one-dimensional Non-linear Hawkes process and (b) depicts the absolute error between the theoretical and estimated kernel}
    \label{Fig:negativekernel}
    \end{figure}
 For the specified parameters, simulation is conducted over a period of $(0,8000]$,  generating a dataset of length $3000$. Despite the theoretical kernel having a separate time and mark function, estimation is performed by approximating the kernel as a combined function of time and mark, following the model given in Equation \ref{Eq:NNModel}. The estimation took $1160$ seconds to converge to the optimal network parameters based on the stopping criteria.  Figure \ref{Fig:negativekernel} displays the theoretical kernel, and the estimated kernel using the NNNH with marks and the corresponding error plot. The kernel and the error plot, depicted in \ref{Fig:negativekernel}, suggest that the NNNH with marks can accurately approximate the given kernel. A comparison is also made with the vanilla NNNH estimation for the test data using the QQ plot presented in Figure \ref{Fig:NNNHqqplot}. From the QQ plot, it is evident that both the NNNH with marks and vanilla NNNH effectively predict the interarrival time for the test dataset.
    
	\subsubsection{Multi-dimensional Hawkes Estimation}

\begin{figure}[ht]
     \centering
     \begin{subfigure}[b]{0.45\textwidth}
        \centering
        \includegraphics[width=\linewidth]{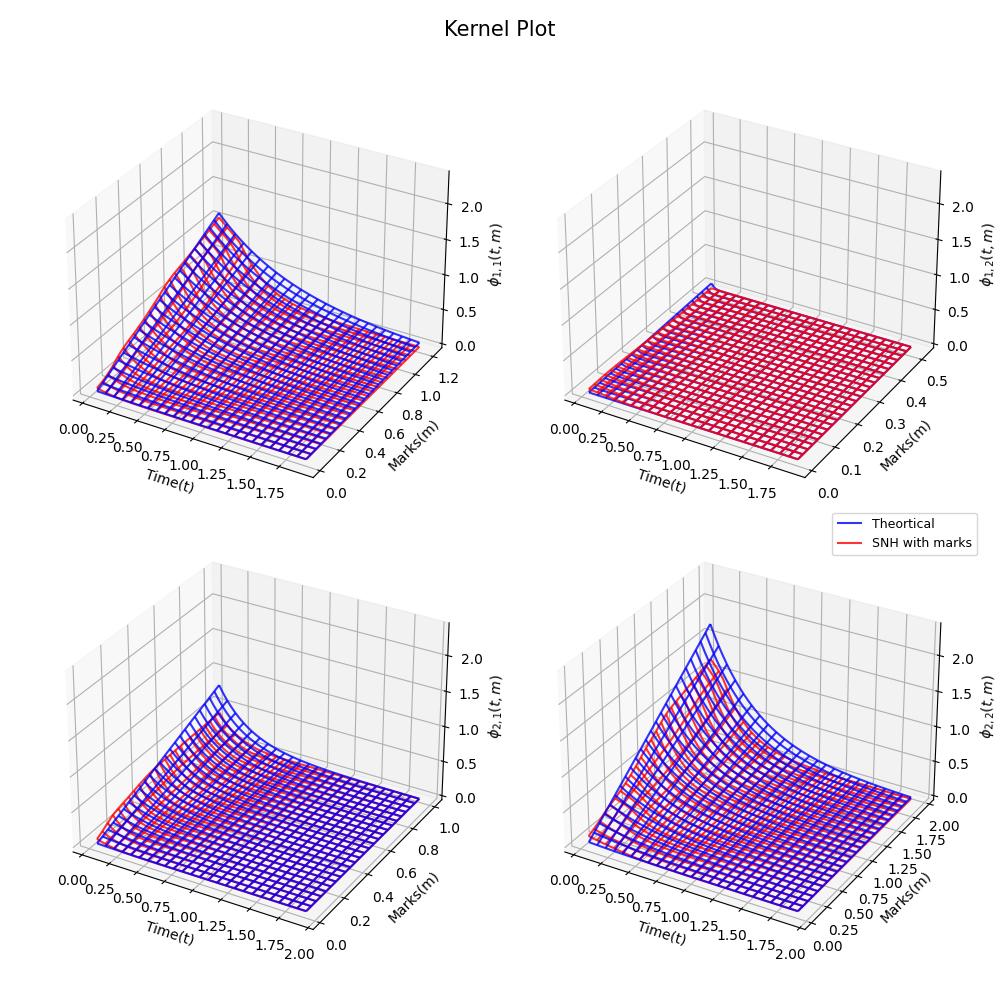}
        \caption{Kernel}\label{Fig:multiKernel}
     \end{subfigure}
     \hfill
     \begin{subfigure}[b]{0.45\textwidth}
         \centering
        \includegraphics[width=\linewidth]{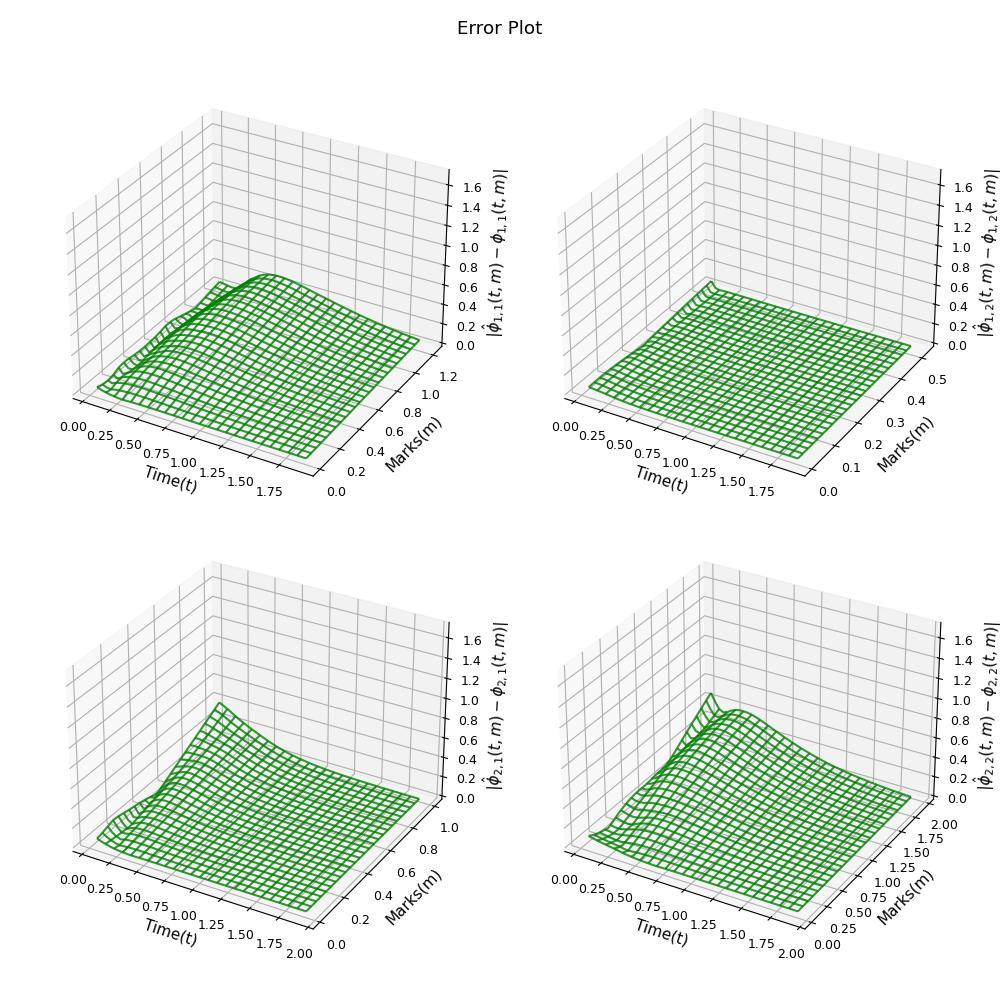}
        \caption{Error}\label{Fig:multiError}
    \end{subfigure}
    \caption{(a) The theoretical and the estimated kernel obtained using SNH with marks for a $2-d$ marked Hawkes process with decoupled kernel function. (b) illustrates the error between the theoretical and estimated kernels}
    \label{Fig:multiExpKer}
    \end{figure}

\begin{figure}[ht]
        \centering
        \includegraphics[scale=0.3]{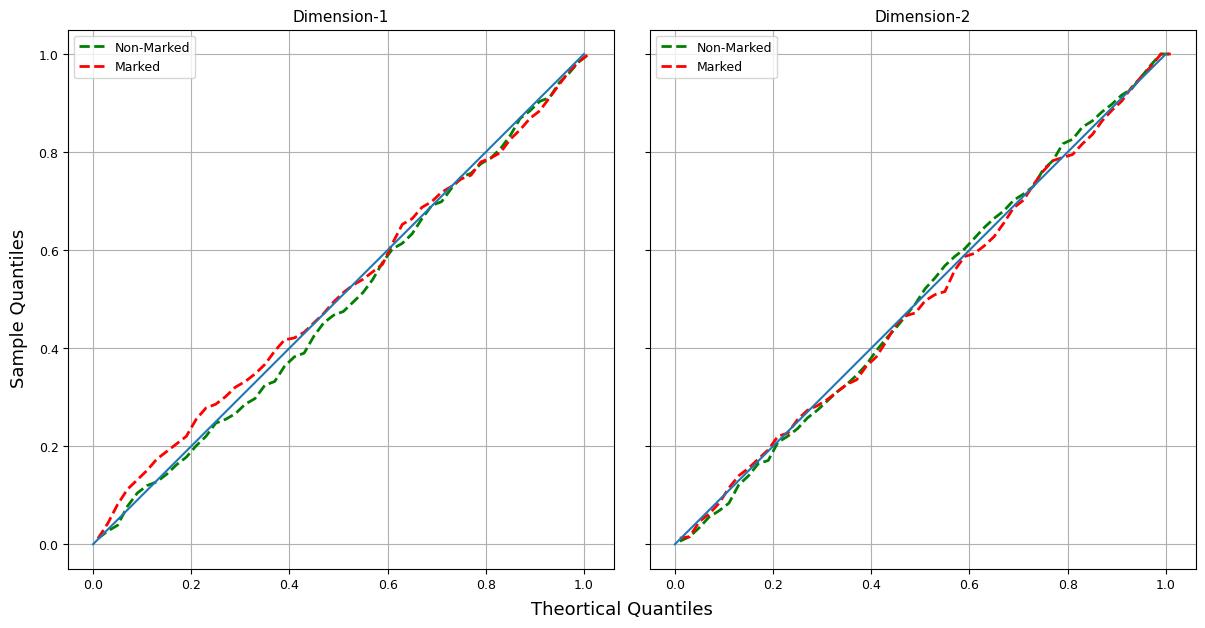}
        \caption{QQ plot demonstrating the precision of prediction for multivariate SNH and SNH with marks estimation.}
        \label{Fig:multiQQplot}
    \end{figure}

   We aim to investigate a $2$-dimensional  marked Hawkes process with distinct kernel functions for temporal and mark components, as defined in the intensity Equation \ref{Eq:multidimMarkedHawkes2}. This specific example, borrowed from \cite{dassios2013exact}, explores a scenario where each kernel possesses its unique marks generating function $f_{dj}(m)$, signifying that each kernel is associated with its own set of marks. The simulation is conducted for a 2 dimensional Hawkes process for a period of $[0,5000)$ with the kernel function, 
    \[ \phi_{dj}(t,m) = \begin{bmatrix}
       me^{-2t} & me^{-50t} \\
       me^{-4.5t} & me^{-3t}     
    \end{bmatrix},\]
     the mark density function,
     \[ f_{dj}(m) = \begin{bmatrix}
        e^{-2.5m} & e^{-6m}\\
        e^{-3m} & e^{-1.5m}     
    \end{bmatrix},\] 
    and base intensity, $\mu= [0.3,0.3]$. The simulation generated a dataset of size of $2082$. The dataset is subjected to the estimation process using multi-dimensional SNH with marks, and the kernels are subsequently obtained. 
    Figure \ref{Fig:multiExpKer}(a) presents a comparison between the estimated kernels and their theoretical counterparts, while Figure \ref{Fig:multiExpKer}(b)  illustrates the absolute error between the true and estimated kernels. These plots collectively affirm that multidimensional SNH with marks has the capacity to recover the theoretical kernel with precision. Furthermore, a QQ plot is provided for both multidimensional SNH with marks and the vanilla SNH in Fig \ref{Fig:multiQQplot}. The QQ plot illustrates that both methods, despite vanilla SNH not considering marks, can predict interarrival times within the given confidence interval. 
	\subsection{Real Data}\label{Subsec:realdata}  
	This section discusses the application of the proposed estimation method in the real-world dataset, where we take the case of cryptocurrency trade in an exchange, specifically the market order book data of Bitcoin-US Dollar(BTC-USD) and Ethereum-US Dollar(ETH-USD) trades in the Binance exchange. The application uses the volume of trade as the mark component, thus analysing the impact of the traded size on the market order arrival rate. This application gives a glimpse of the ability of our method to model real-world events with attached marked components. 
	
	\subsubsection{Cryptocurrency Dataset} \label{Subsubsec:findata}
	\paragraph{Data and Preprocessing:}
	   We utilize the high-frequency dataset from cryptocurrency trading involving two of the most commonly traded cryptocurrencies- Bitcoin (BTC) and Ethereum (ETH). The dataset consists of timestamps of buy and sell market orders for Bitcoin-US Dollars(BTC-USD) and Ethereum-US Dollars (ETH-USD) pairs on the Binance exchange. The Binance exchange was selected because it is a major exchange for various cryptocurrencies with high trading volumes. The dataset details arrival timestamps, volume, price, buyer ID and seller ID for each market order. The time horizon considered is from 16th July 2022, from $02:00:00$ UTC to $02:30:00$ UTC. The selection of Bitcoin and Ethereum currencies is based on the fact that they are the most commonly and heavily traded cryptocurrencies in the global cryptocurrency market.
    
    For the cryptocurrency dataset, we adopt the volume of trade as the mark component in the marked Hawkes process. This aligns with the approach of \cite{chavez2012high}, \cite{fauth2012modeling}, \cite{rambaldi2017role} and \cite{fabre2024neural}, which explore the impact of volume of trade on the clustering of market order. This justifies the use of volume as the mark component. In this context, the trade volume represents the quantity of bitcoins (or Ethereum) traded.  
 
    Before applying the estimation method, some preprocessing is done to the dataset. There are instances in which a single sell/buy market order is fulfilled by multiple buyers (sellers) in the market. These orders are considered as a single sell (buy) market order, combining their respective trading volumes. Table \ref{Tab:cryptodata} shows the descriptive statistics of the dataset after preprocessing. The table shows that the trade volume for ETH-USD currency (both buy and sell) is 10 times greater than the BTC-USD currency trade despite the higher number of BTC-USD trades. This discrepancy is primarily due to the higher price of Bitcoin compared to Ethereum, resulting in a greater quantity (volume) of ETH-USD in a single transaction compared to BTC-USD.
        
\begin{figure}[ht]
     \centering
        \includegraphics[scale=0.2,width=\linewidth]{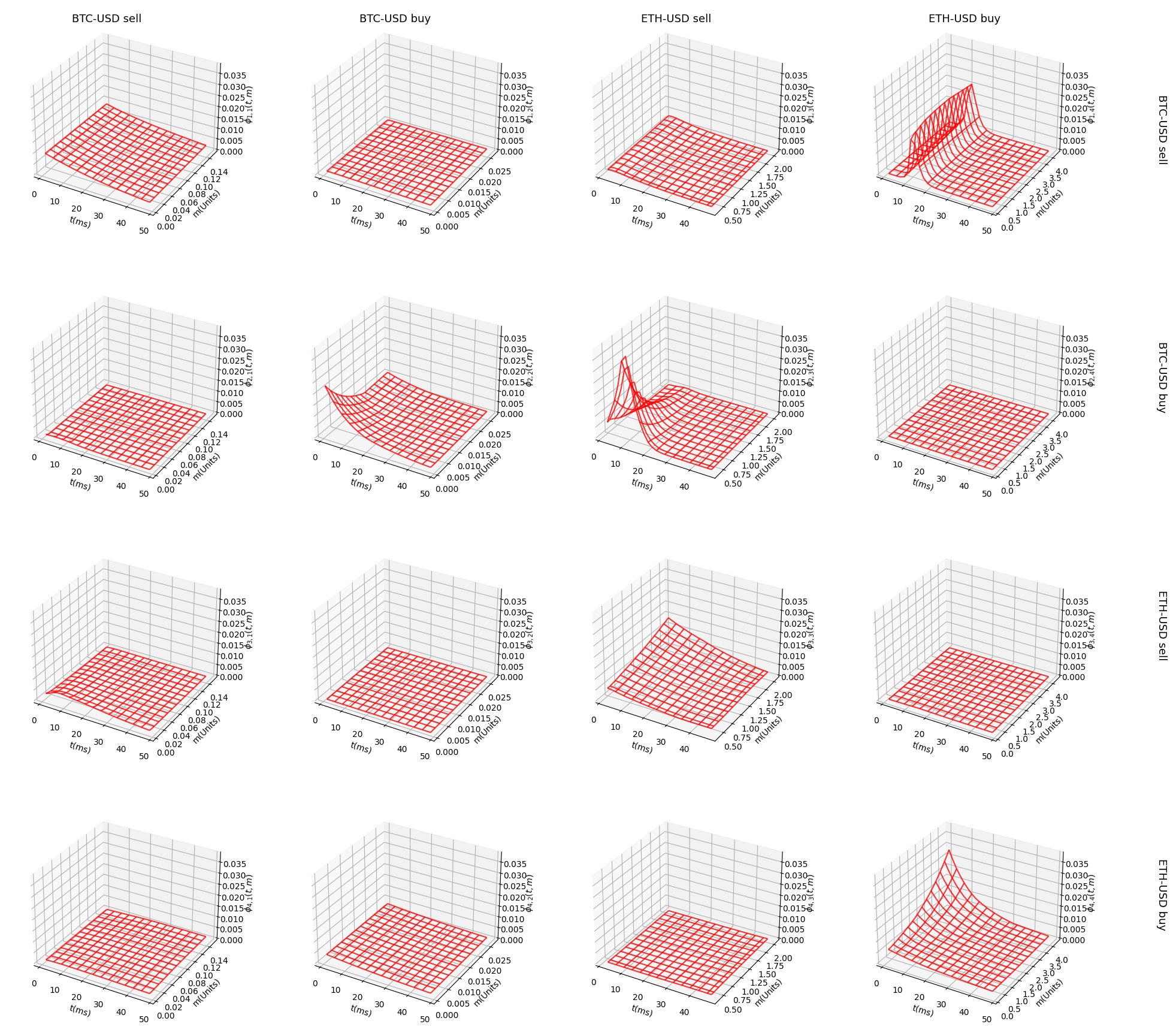}
    \caption{Kernels for the given cryptocurrency dataset estimated using SNH with marks}
    \label{Fig:binanceKer}
    \end{figure}

\begin{figure}[ht]
        \centering
        \includegraphics[scale=0.3]{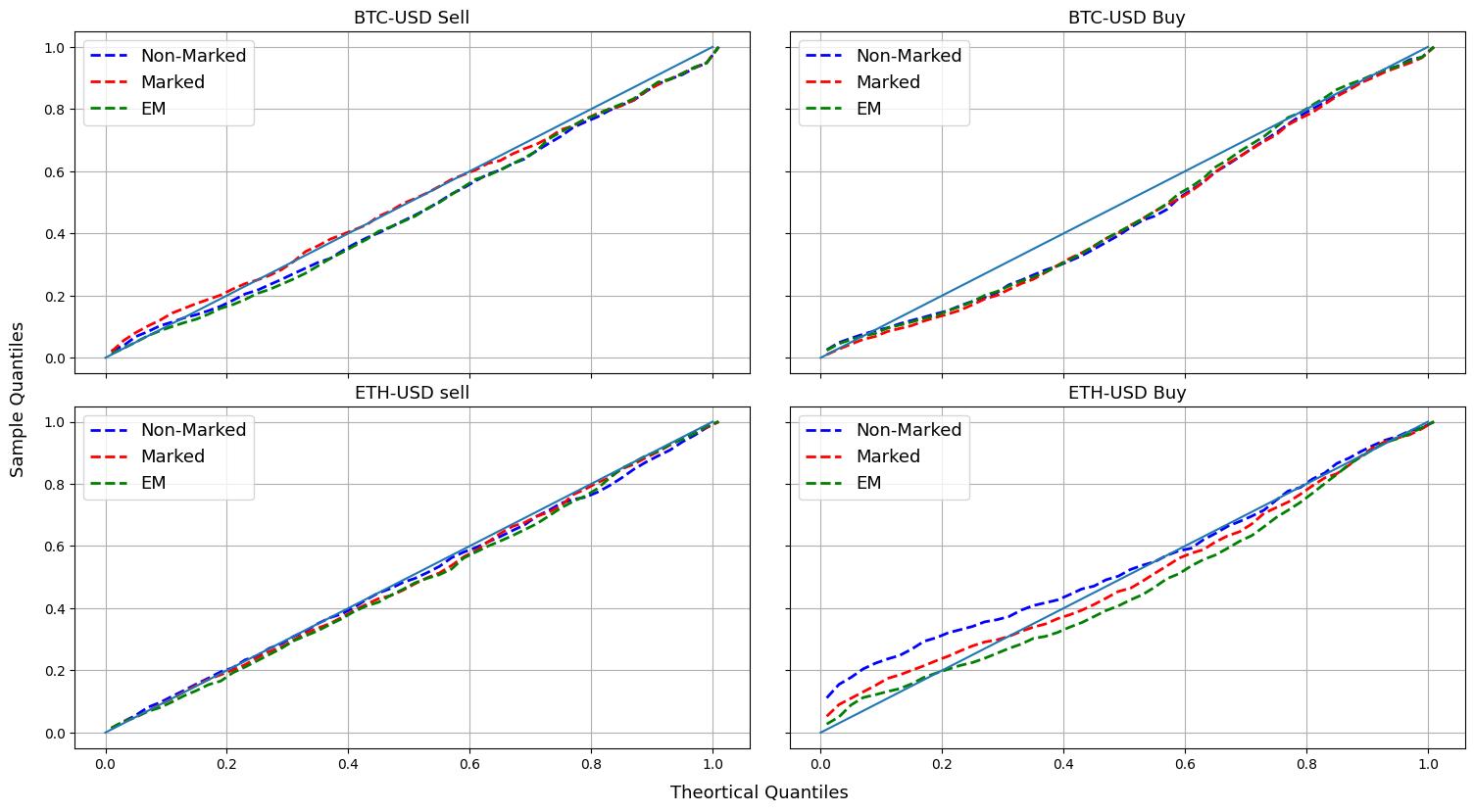}
        \caption{QQ plot comparing the performance of SNH with marks, vanilla SNH and EM estimation methods.}
        \label{Fig:binanceqqplot}
    \end{figure}

   \begin{table}[ht]
    \caption{Descriptive statistics of the cryptocurrency market orders}
    \begin{center}
    \begin{tabular}{|c|c|c|c|}
        \hline
           Cryptocurrency & Trade-Type & Number of Events & Volume of trade  \\ \hline
         BTC-USD & sell & $7512$ &  $463.23$  \\ \hline
         BTC-USD & buy & $9046$ &  $734.72$  \\ \hline
         ETH-USD & sell & $6989$ & $9139.5$ \\ \hline
         ETH-USD & buy & $6270$ & $8101.3$ \\ \hline
         Total   & & $30388$ & $18438.75$\\ \hline
    \end{tabular}
    \end{center}
    \label{Tab:cryptodata}
	\end{table}
	\paragraph{Result and Summary:}
    Figure \ref{Fig:binanceKer} illustrates the kernels, $\phi_{dj}(t,m)$ obtained after applying SNH with marks model to the cryptocurrency dataset. From Figure \ref{Fig:binanceKer} the following can be observed regarding the trade arrivals of BTC-USD and ETH-USD trades:
    \begin{itemize}
        \item  All the cryptocurrency trades(both buy and sell) exhibit self-excitatory behaviour but the amount of self-excitation varies for each currency.  The self-excitation remains constant with the volume of trade for sell BTC-USD trade but decreases with the volume for buy BTC-USD trades. While, for buy and sell ETH-USD trades, the intensity of trade arrival due to its trades increases with the volume of trade.
        \item Cross excitatory behaviour is observed in sell BTC-USD and buy ETH-USD. The occurrence of trades in buy ETH-USD triggers sell BTC-USD trades. This triggering intensity gradually increases with an increase in trading volume.  
        \item Trades in sell ETH-USD cross excite more trades in buy BTC-USD. This excitation is observed only for a small volume of  ETH-USD trade and diminishes with an increasing volume.
    \end{itemize}
    The above observations give the microstructure in a cryptocurrency market, i.e. Bitcoin trades (buy/sell) are influenced by Ethereum (sell/buy) trades, but not the other way around. There is also a clear dependency on the volume of trade with the intensity of trade. This statement will also be supported by the QQ plot given in \ref{Fig:binanceqqplot}.

    The QQ plot compares the performance of SNH with marks, vanilla SNH and EM.{\footnote{ EM is Expectation Maximization algorithm for Hawkes process estimation proposed by \cite{lewis2011nonparametric}.}} The QQ pots for sell BTC-USD and buy ETH-USD indicate a volume dependency, suggesting that the volume of trade plays a pivotal role in the arrival of the next trade, while there is no observable dependency for the volume of trade in the other two dimensions (buy BTC-USD and sell ETH-USD).
    
\begin{figure}[ht]
     \centering
     \begin{subfigure}[b]{0.45\textwidth}
        \centering
        \includegraphics[width=\linewidth]{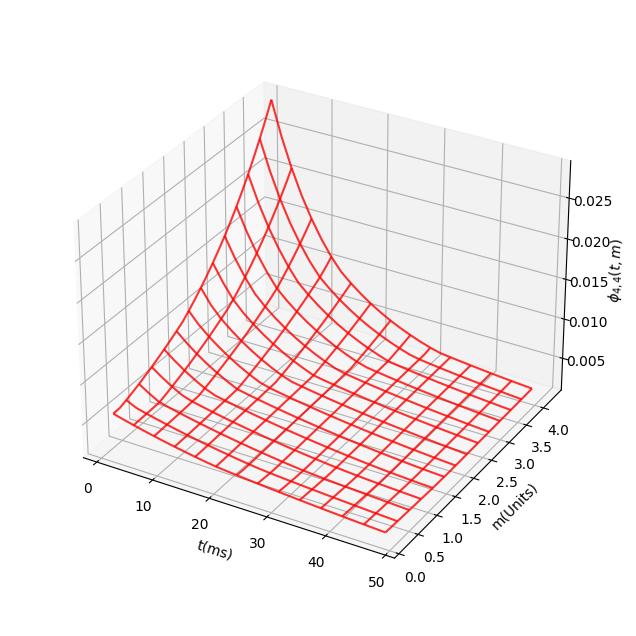}
        \caption{SNH with marks}\label{Fig:binance3d}
     \end{subfigure}
     \hfill
     \begin{subfigure}[b]{0.45\textwidth}
         \centering
        \includegraphics[width=\linewidth]{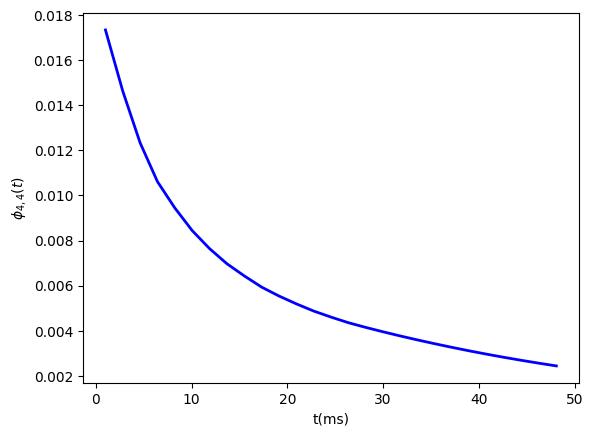}
        \caption{Vanilla SNH}\label{Fig:binance2d}
    \end{subfigure}
    \caption{Estimated self-excitatory kernel of buy ETH-USD indicating the difference with marks and without marks estimation }
    \label{Fig:markComMarkNomark}
    \end{figure}

    In the context of market microstructure,  understanding market dynamics in relation to various factors is crucial. The application of SNH with marks to the cryptocurrency dataset contributes to understanding the change in trade intensity concerning both time and volume. Figure \ref{Fig:markComMarkNomark} illustrates the self-excitatory kernel of ETH-USD estimated using SNH with marks ($\phi_{4,4}(t,m)$ in Figure \ref{Fig:binanceKer}) and using Vanilla SNH. Figure \ref{Fig:binance3d} shows a clear mark dependency for the kernel with the intensity increasing as the volume of trade increases. In contrast, kernels from vanilla SNH (Figure \ref{Fig:binance2d}) don't consider the volume, providing an average intensity across all the volumes at a given time $t$.  Thus using SNH with marks one can deduce the dependency of the volume on kernel intensity, a feature absent in vanilla SNH.

    \subsection{Prediction of Marked Hawkes process}
    The prediction of marked Hawkes is closely tied to the simulation process. The key quantities of interest for prediction would be the time to the next event (interarrival time) and the probability of an event occurring within a  given interval of time (exponential of the integrated intensity). These quantities can be derived if the conditional intensity is known, and the functional form of the conditional intensity is obtained through the discussed estimation methods. If the structure of the conditional intensity is known, conditional intensity can be obtained with any arbitrary history of time and mark. Thus it is straightforward to see prediction as an application and extension of the preceding procedures. A step-by-step detail of the prediction process is given as,
    \begin{enumerate}
        \item Select a time period of the dataset $(0,A]$ to estimate the marked Hawkes process, thereby obtaining the functional form of conditional intensity function.
        \item Simulate the process beyond $(0,A]$ using Algorithm \ref{alg:ThinningMarked} leveraging the known structure of the conditional intensity function and the known history $H_{A}$.
        \item Extract from the simulation the quantity required to predict (interarrival time or probability of arrival).
        \item Repeat steps 2 and 3 a sufficient number of times to achieve the desired precision for the prediction quantity.
        \item The output is the empirical distribution of the quantity obtained from multiple successive simulations.
    \end{enumerate}

	\section{Conclusion}\label{Sec:conclusion}
	
	This paper introduces a novel non-parametric estimation technique tailored for the marked Hawkes process. This is the first known approach capable of non-parametrically estimating the kernel of the Hawkes process as a joint function of time and marks. Drawing inspiration from the vanilla SNH method and the NNNH method, the proposed model employs a two-layer feed-forward neural network to approximate the Hawkes kernel, treating the kernel as a joint function of event times and their associated marks. This model is applied to the log-likelihood function and the optimal neural network parameters are derived using batch SGD combined with Adam. The use of batch SGD  renders the method suitable for online learning applications, facilitating the updation of network parameters based on the arrival of new data points. 

    Here we explain two approaches SNH with marks- tailored for the linear marked Hawkes process and NNNH with marks- for the non-linear marked Hawkes process. Even though the NNNH with marks possesses the capability to model linear Hawkes models, we limit its use to scenarios suspected of negative dependencies as clarified in Remark \ref{rem:Comparison}. We apply the estimation methods to synthetic and cryptocurrency datasets. We compare all the results obtained with the corresponding estimation methods without marks, and QQ plots are obtained. Although QQ plots for estimation with marks and without marks appear unique (in some datasets), using estimation with marks proves advantageous when understanding the dependency of event arrivals on marks is crucial. In the cryptocurrency setting, estimation with marks reveals the relationship of the market order arrival intensity with the volume of the trade apart in addition to the history of trade arrivals. This provides a comprehensive insight into the dynamics of the cryptocurrency market microstructure.
    
   While the proposed methods prove robust in estimating Hawkes process with marks, determining the appropriate mark component in real-life applications remains a relevant factor, which can be taken as a future area of focus. Additionally, incorporating spatial coordinates as marked components could be a  potential focus for future directions. Including spatial elements in the model has the potential to enhance our understanding and prediction capabilities in earthquake modelling and related fields.

\bibliography{paper3}
\bibliographystyle{plainnat}

\medskip
	\appendix
    \section{Approximation of the Log-likelihood Function}\label{App:ApprLL}
    The ground intensity part of the log-likelihood function, $\mathcal{L}_{\lambda}$ given in the Equation \ref{Eq:LLLambda} can be partitioned into two components: the \textit{log intensity} part, $\log(\lambda^{g}_d(t^d_n))$ and the \textit{integrated intensity} part, $\int_0^{T} \lambda^{g}_d(s) ds$.
    
    For both the introduced models- SNH with marks and NNNH with marks, it is necessary to obtain both the integrated intensity and log intensity functions to obtain the gradients from Equation \ref{Eq:LLGrad}.  Although obtaining the log intensity part is straightforward, the integrated intensity part is not so direct for both the SNH with marks and NNNH with marks. 
    \subsection{Integrated Intensity of SNH with marks}\label{App:IntegrSNH with marks}
    For the SNH with marks, as the intensity function, $\lambda_d(t)$ is linear, it is written as the arithmetic combination of base intensity, $\mu$ (taking the baseline intensity as constant) and the kernel functions, $\phi_{dj}(t),  \forall 1 \le j \le D$. According to \cite{joseph2022shallow}, the integrated intensity function for a linear Hawkes process without marks can be written as,
    \begin{equation*}
        \int_0^{T} \lambda_d(s) ds = \sum_{t_n^d \in \boldcal{H} } \int_{t_{n-1}^d}^{{t_{n}^d}} \mu_d ds+ \sum_{j=1}^D \sum_{t_n^j \in \boldcal{H}} \int_{0}^{T-t^j_n} \phi_{dj}(s) ds.
    \end{equation*}
    The above equation can be modified for a marked Hawkes process. Thus, the integrated ground intensity with modification in  $\phi_{dj}$ to accommodate marks is written as
        \begin{equation}
        \label{Eq:linKerMark}
        \int_0^{T} \lambda^{g}_d(s) ds = \sum_{t_n^d \in \boldcal{H}} \int_{t_{n-1}^d}^{{t_{n}^d}} \mu_d ds+ \sum_{j=1}^D \sum_{\{t_n^d,m_n^d\} \in \boldcal{H}} \int_{0}^{T-t^j_n} \phi_{dj}(s,m_n^j) ds.
    \end{equation}
    For the SNH with marks, the above kernel can be approximated according to Equation \ref{Eq:SNHmodel},
    \begin{equation*}
		\hat{\phi}_{dj}(s,m) = \text{exp}\left(b_2 + \sum_{i=1}^{P}a^i_3\text{max}\left( a^i_1 s + a^i_2 m +b^i_1, 0\right)\right).
	\end{equation*}
    Each of the integrated kernel function $\int_0^{T-t^j_n} \hat{\phi}_{dj}(s,m_i^j) ds $ in Equation \ref{Eq:linKerMark} involves integrating over the neural network model given in Equation \ref{Eq:SNHmodel}. As the model consists of the RELU function, it is necessary to integrate over  \textit{max} function for each of the $i$th neuron, $\forall 1 \le i \le P$.  This involves obtaining the \textit{zero-crossings} and integrating over non-zero paths of each $i$th neuron over the interval $[0,T-t^j_n)$. For the $i$th neuron, the \textit{zero-crossing}, $y_i$ is given by,
    \begin{eqnarray*}
        a^i_1 y_i + a^i_2 m_n^j +b^i_1 = 0 \\
        y_i=-\frac{a^i_2 m_n^j+b^i_1}{a^i_1}.
    \end{eqnarray*}
    Let $\mathcal{Y}= (y_1,\ldots, y_P)$ be the \textit{zero crossings} across all the $P$ neurons. Arranging the $\mathcal{Y}$ in the increasing order and within the bounds of $[0,T-t^j_n)$ given by $\mathcal{Y^*} = (0, \, y^*_1, \ldots , y^*_u, \, T-t^j_n)$, where $u\le P$. Thus the integrated kernel function across each of $[0,T-t^j_n)$ is given as,
    \begin{equation}\label{Eq:IntegratedSNH}
        \int_0^{T-t^j_n} \hat{\phi}_{dj}(s,m_n^j) dt = \int_0^{y^*_1} \hat{\phi}_{dj}(s,m_n^j) ds +  \int_{y^*_1}^{y^*_2} \hat{\phi}_{dj}(s,m_n^j) ds + \ldots + \int_{y^*_u}^{T-t^j_n} \hat{\phi}_{dj}(s,m_n^j) ds.
    \end{equation}
    The obtained approximated integrated intensity function is substituted to the $\mathcal{L}_{\lambda}$ in Equation \ref{Eq:LLLambda} and is maximized to obtain the network parameters.
    \subsection{Integrated Intensity of NNNH with marks} \label{Sec:IntegrNNNH with marks}
    For the NNNH with marks method, the calculation of the integrated intensity function is complex compared to SNH with marks due to the presence of two non-linear functions ($\max$ functions). The integrated intensity function, $\int_0^{T} \lambda^{g}_d(s)ds$ can be written as,
    
    \begin{eqnarray*}
        \int_0^{T} \lambda^{g}_d(s)ds &=&  \sum_{\{t^d_n,m^d_n\} \in \boldcal{H}}\int_{t_{n-1}^d} ^ {t_{n}^d} \lambda^{g}_d(s) ds,
    \end{eqnarray*}
     the sum of all distinct arrival times, as it is more convenient to calculate the intensity function across all distinct intervals.  From the non-linear marked Hawkes intensity function given in Equation \ref{Eq:NonLinearHawkes}, $\lambda^{g}_d(s)$ across each distinct interval is,
     \begin{eqnarray*}
        \int_{t_{n-1}^d} ^ {t_{n}^d} \lambda^{g}_d(s) ds &=& \int_{t_{n-1}^d} ^ {t_{n}^d} \max \left( \left[\mu_d + \sum_{j=1}^D\sum_{t^j_k<s} \phi_{dj}(s-t^j_k,m^j_k)\right],0 \right) ds, \\
        &=& \int_{t_{n-1}^d} ^ {t_{n}^d} \max \left( \left[\mu_d + \sum_{j=1}^D\sum_{t^j_k<s} b_2 + \sum_{i=1}^{P} a^i_3 \max \left( a^i_1 (s-t^j_k) + a^i_2 m^j_k +b^i_1, 0\right) \right],0 \right) ds,
    \end{eqnarray*}
    the second equality substitutes neural network model from Equation \ref{Eq:NNNHmodel} in to $\phi_{dj}(s-t^j_k,m^j_k)$.  For finding out  $\int_{t_{n-1}^d} ^ {t_{n}^d} \lambda^{g}_d(s) ds$, it is necessary to integrate the function over non-zero paths. This is done by determining the \textit{zero-crossings} across each $[{t_{n-1}^d}, {t_{n}^d}]$ pairs. For finding out \textit{zero-crossings} for an given interval first, we have to find the \textit{zero-crossings} for the inner $\max$ function layer and then the outer $\max$ layer.  Let $\mathcal{Y}= (y_1,\ldots, y_{P})$ be the \textit{zero-crossings} for the entire neural network with $P$ neurons (inner $\max$ layer), $y_i$ for each $i$th neuron is given by,
    \begin{eqnarray*}
        a^i_1 (y_i-t^j_k) + a^i_2 m^j_k +b^i_1 = 0, \\
        y_i=\frac{a^i_1 t^j_k-a^i_2 m^j_k-b^i_1}{a^i_1}.    
    \end{eqnarray*}
    Thus for each of the $\int_{t_{n-1}^d} ^ {t_{n}^d} \lambda^{g}_d(s) ds$, there will be $ F = PD l(\mathcal{H}_{s^-})$ ($l(\mathcal{H}_{s^-})$ denotes the length of history up to $s$) number of zero crossings for the inner $\max$ layer. Arranging $\mathcal{Y}$ in chronological order and in the bounds of  $\left[{t_{n-1}^d}, {t_{n}^d}\right]$ given by 
    $\mathcal{Y}'= (t_{n-1}^d, \, y_1, \ldots, y_{F},\, t_{n}^d)$.
    Next, we found out the \textit{zero-crossings} across the outer $\max$ layer. This is done by finding the \textit{zero-crossings} in each of the pairs in $\mathcal{Y}'$. For each of the pair in $\mathcal{Y}'$, there will be at most one \textit{zero-crossings}, denoted by $y^*_i$. Thus the modified $\mathcal{Y}'$ with all the \textit{zero-crossings} across the interval $\left[{t_{n-1}^d}, {t_{n}^d}\right]$, $\mathcal{Y}^* = (t_{n-1}^d, \, y^*_1, \, y_1,\ldots, \, y_F, \,t_{n}^{d*} , \,t_{n}^d)$ ($\mathcal{Y}^* $ includes \textit{zero-crossings} of both inner and outer $\max$ function)

     Finally, the integrated intensity, $\int_{t_{n-1}^d} ^ {t_{n}^d} \lambda^{g}_d(s) ds$ is given by,
    \begin{equation}
        \int_{\mathcal{Y}^*} \lambda^{g}_d(s) ds = \int_{t^d_{n-1}}^{y^*_1} \lambda^g_d(s)ds +  \int_{y^*_1}^{y_1} \lambda^g_d(s)ds +  \ldots  +  \int_{y_F}^{t^{d*}_{n}} \lambda^g_d(s)ds + \int_{t^{d*}_{n}}^{t^d_{n}} \lambda^g_d(s)ds,
    \end{equation}
    where $\lambda^g_d(s)=\mu_d + \sum_{j=1}^D \sum_{t_i^j < T} \int_{t^d_n}^{t^d_{n-1}} \phi_{dj}(s,m)$.

\end{document}